\documentclass[11pt,a4paper]{article}

\usepackage[utf8]{inputenc}
\usepackage[T1]{fontenc}
\usepackage{amsmath,amssymb,amsfonts}
\usepackage{graphicx}
\usepackage{hyperref}
\usepackage{geometry}
\usepackage{authblk}
\usepackage{cite}

\usepackage{tikz}
\usetikzlibrary{fit, backgrounds}
\tikzstyle{mybox} = [draw=black, very thick, rectangle, rounded corners, inner ysep=5pt, inner xsep=5pt]

\usepackage{siunitx}
\usepackage{enumitem}

\usepackage{amsmath}
\usepackage{authblk}
\usepackage{lettrine}
\usepackage{algpseudocode} 
\usepackage{amssymb}  
\usepackage{amsfonts}
\usepackage{siunitx}
\usepackage{graphicx,subfigure}
\usepackage{enumitem}

\hypersetup{
  colorlinks = true,
  linkcolor = blue,
  citecolor = blue,
  urlcolor = blue
}

\title{\textbf{Sparse and nonparametric estimation of equations governing dynamical systems with applications to biology}}

\author[1]{G. Pillonetto}
\author[2]{A. Giaretta}
\author[3]{A. Aravkin}
\author[1]{M. Bisiacco}
\author[4]{T. Elston}
\affil[1]{Department of Information Engineering, University of Padova, Padova (Italy)}
\affil[2]{Department of Pathology, Cambridge University, Cambridge (UK)}
\affil[3]{Department of Applied Mathematics, University of Washington, Seattle (USA)}
\affil[4]{Department of Pharmacology, University of North Carolina at Chapel Hill (USA)}

\date{} 

\begin{document}
\maketitle  

\begin{abstract}
Data-driven discovery of model equations is a powerful approach for understanding the behavior of dynamical systems in many scientific fields.  
In particular, the ability to learn mathematical models from data would benefit systems biology, where the complex nature of these systems often makes a bottom up approach to modeling unfeasible. 
In recent years, sparse estimation techniques have gained prominence in system identification, primarily using parametric paradigms to efficiently capture system dynamics with minimal model complexity. In particular, the Sindy algorithm  has successfully used sparsity to estimate nonlinear systems by extracting from a library of functions only a few key terms needed to capture the dynamics of these systems. However, parametric models often fall short in accurately representing certain nonlinearities inherent in complex systems. To address this limitation, we introduce a novel framework that integrates sparse parametric estimation with nonparametric techniques.
It captures nonlinearities that Sindy cannot describe without requiring a priori information about their functional form. That is, without expanding the library of functions to include the one that is trying to be discovered. We illustrate our approach on several examples related to estimation of complex biological phenomena.
\end{abstract}

\section{Introduction}
Data-driven discovery of governing model equations is a powerful approach to understanding the behavior of natural phenomena. This is particularly true for dynamical systems described by differential equations, where estimation problems fall under the umbrella of system identification \cite{Astrom71,Soderstrom,Ljung:99}. 
Recently, there has been a convergence of system identification and machine learning methods \cite{Tibshirani:96,SpringerRegBook2022,PillonettoPNAS}.  This synergy uses regularization theory to construct minimal models using combinations of terms from predefined candidate libraries chosen to reproduce empirical data. The principle of parsimony remains central to this approach, achieved by imposing penalties on solution complexity \cite{Bai2019,Rosasco2013,Smith2014,Stoddard2017}. The emerging field of physics-informed machine learning takes this concept further by incorporating domain-specific physical knowledge into regularizers for dynamical systems models \cite{Champion2020,Karniadakis2021,nghiem2023,Raissi2019}. Several techniques use norms as penalties, such as those induced by kernels \cite{Scholkopf01b}. A kernel is a symmetric, positive-definite map that defines a space of functions and also encodes their regularity properties. It controls how smooth or complex the functions in that set can be. For example, the spline kernel measures the complexity of a function  through the integral of the squared derivative \cite{Wahba:90}. So, if two functions describe the data in the same way, the spline estimator
selects the more regular one, i.e. with smaller energy of derivative. 
Kernels also give rise 
to so-called nonparametric estimators, such as regularization networks and support vector machines, since they have a very large (potentially infinite) number of parameters \cite{Tibshirani2001,PC08,Daub2004}. \\

\begin{figure*}
	\begin{center}
			{ \includegraphics[scale=0.5]{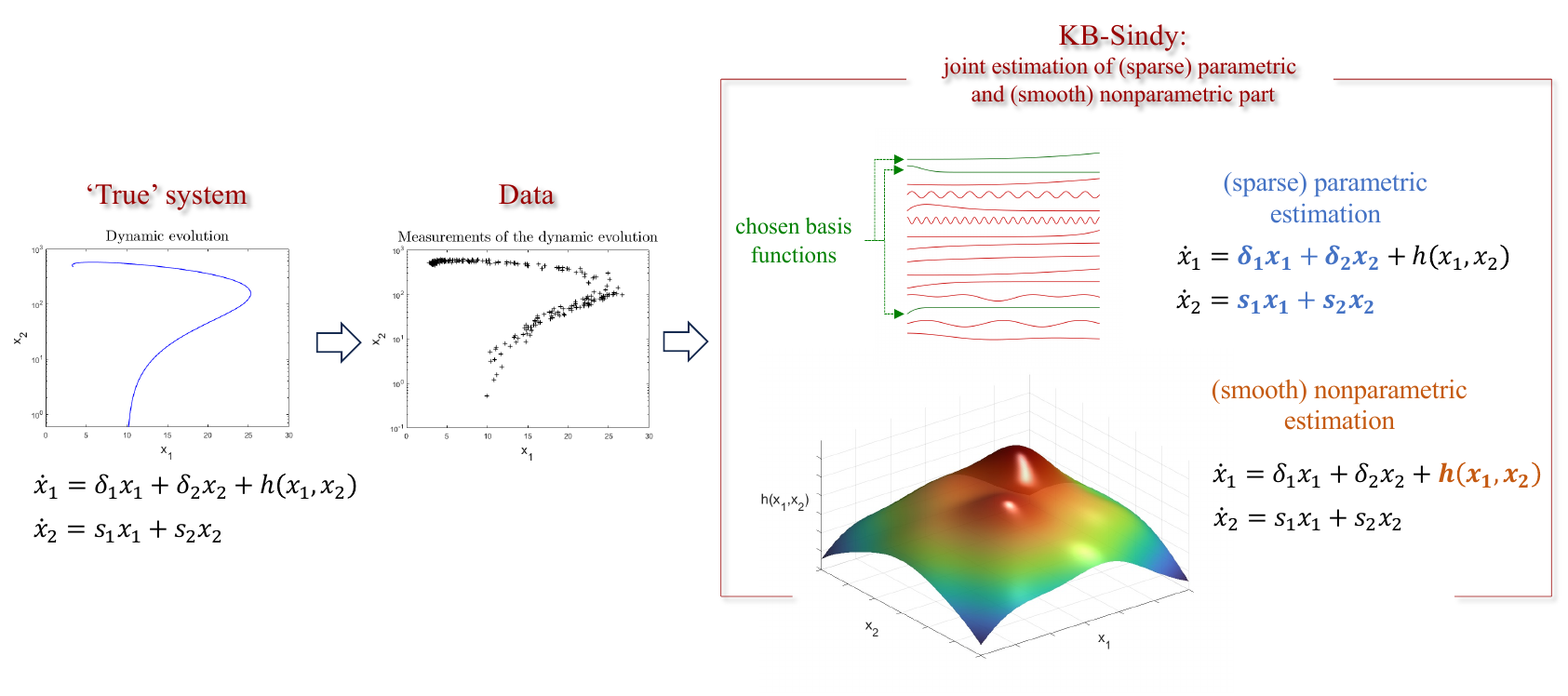}} 
		\caption{KB-Sindy overview. An experiment provides noisy time series data for a dynamical system.  KB-Sindy models the system as the sum of two parts. The first is parametric and described in terms of a prescribed set of basis functions. The complexity of the parametric component is controlled by sparsity information.  The function  $h$ takes into account terms  difficult to approximate using the basis functions and is therefore estimated in a nonparametric way. The complexity of $h$ is controlled by a machine learning approach based on a prescribed kernel, which only encodes information about the regularity of $h$. 
        }  \label{FigIntro}
	\end{center}
\end{figure*}

Another learning method is the Sindy algorithm \cite{Brunton2016,Champion2019}. It uses sparsity-promoting regularization and sequential least squares to select a minimal set of terms from a function library that capture the system's dynamics.
An advantage of Sindy over kernel methods is that it returns models that are interpretable, providing insight into the mechanisms that govern the system under consideration.
However, sparse estimation strategies such as the Sindy algorithm rely primarily on parametric models, e.g. the truncated Volterra series \cite{Boyd1985,Stoddard2017}, which may struggle to accurately capture certain nonlinearities inherent in complex systems. For example, certain biochemical rate laws, such as Hill kinetics,  are difficult to capture by monomials \cite{Elowitz2000,Giaretta2020}. Furthermore, many natural systems are subject to stochastic dynamics that can qualitatively change the systems behavior   \cite{Ai2003,Vasseur2004,Bjornstad2001,Ruokolainen2009}.
To allow for these possibilities, we present an extension of Sindy that is particularly suited to problems where the system structure is not fully known a priori, but partial information about its nature is available. It integrates sparse parametric estimation with nonparametric techniques by combining Sindy and kernel methods. This hybrid approach, which we call Kernel-based Sindy (KB-Sindy), exploits the strengths of both methods and allows for a more comprehensive and flexible framework for learning model equations from data. Our approach captures nonlinear relationships that cannot be parsimoniously described by Sindy in the absence
of strong a priori information about their functional form. That is, without adding to the library of functions the nonlinear interactions that are 
trying to be discovered. KB-Sindy can also handle stochastic dynamics because it is not constrained by a predetermined functional form. Fig. \ref{FigIntro} provides an overview of KB-Sindy: the system is modeled as the sum of two parts, one parametric with complexity controlled by sparsity constraints and the second nonparametric relying on a kernel approach. This results in an extension of both paradigms, and provides a framework to infer nonlinear models that neither Sindy nor kernel-based methods can accurately capture.\\

Systems biology is an ideal domain for data-driven modeling. It is characterized by complex dynamics, but often incomplete information about the molecular components of the system under consideration \cite{AlonUri2020,Dupont1991}. Therefore, deriving models from biological knowledge alone is challenging.   There is a need for methods that extract governing equations directly from data without imposing structural assumptions while simultaneously allowing  models to be interpretable in terms of biological processes. KB-Sindy is an effective tool to achieve these goals for models based on ordinary differential equations (ODEs) and partial differential equations (PDEs). We successfully apply our method to a diverse set of examples, including the Lorenz equations, gene regulation, spatio-temporal calcium signaling and the logistic map.

\section{Kernel-based Sindy}
Before presenting Kernel-based Sindy (KB-Sindy), we briefly review the Sindy algorithm \cite{Brunton2016}.

\subsection*{Sindy}
A dynamical system is described by a set of coupled ODEs:
\begin{equation}\label{StateMod}
\dot{x}(t) = f(x(t))
\end{equation}
where $x(t) \in \mathbb{R}^n$ is the state of the system at time $t$ and the vector $\dot{x}(t)$ contains the time derivatives of the variables.
The vector function  $f: \mathbb{R}^n \rightarrow \mathbb{R}^n$ is assumed to be unknown, and the goal of Sindy is to infer the form of $f$ from time series data.\\
The main assumptions introduced in \cite{Brunton2016} are that measurements of the state $x$ are available at times $t_1,t_2,\ldots,t_m$ 
together with noisy estimates of the corresponding $\dot{x}$. In addition,  
it assumed that the system $f$ can be written in terms of the expansion
\begin{equation}\label{FirstModelForf}
f(x)= \sum_{i=1}^p \Xi_i \phi_i(x),
\end{equation}
where the $\phi_i$'s form a basis set. 
This reduces the problem to estimating the expansion coefficients $\Xi_i$.
The problem can be formulated in matrix-vector form by
introducing the regression matrix $\Theta$, 
whose columns contain candidate functions from the basis set.
A popular choice is to use a truncated Volterra series in which case the basis functions are monomials up to order $r$ \cite{Boyd1985,Stoddard2017}.  
For instance, if $r=2$ and the state dimension is $n=2$, one obtains $p=5$ and the $i$th row of $\Theta$ becomes
\begin{equation}\label{MonomialEx}
\Theta_i=\left[\begin{array}{ccccc} x_1(t_i) & x_2(t_i)  &  x^2_1(t_i) & x^2_2(t_i) & x_1(t_i)x_2(t_i) \end{array}\right].
\end{equation}
To simplify the presentation, we focus on the estimation of a single component $f_i$ and
obtain 
\begin{equation}\label{MeasModY}
y = \Theta \Xi + e
\end{equation}
where the column vector $y$ contains 
estimates of
the derivatives of $x_i(t)$ at times $t_1,t_2,\ldots,t_m$ and  $e$ represents an additive noise term.\\

In many systems only a few of the columns (monomials) are present in the model equations.
This feature is used by Sindy to control model complexity and learn interpretable models.
The governing equation is estimated from $y$ by sequential least squares where, at any iteration,
the sparsity parameter $\lambda$ determines the number of components of $\Xi$ that are set to zero. 
We summarize the Sindy method and approaches to estimate $\lambda$ in {\bf Materials and Methods}.

\subsection*{KB-Sindy}

We now consider a more complex situation, where only part of $f$ is described by a limited number 
of basis functions. The remainder of $f$ 
contains nonlinearities $h$ 
which are difficult to approximate with a small number of monomials. 
In fact, a limitation of the Sindy method is that the number $p$ of basis functions grows rapidly with the state space dimension $n$ (many monomials are required even for relatively small order $r$ due to the curse of dimensionality).
Therefore, to address these issues, in our method
we assume that $f$ can be written as the sum of two terms:
\begin{equation}\label{SecondStateMod}
\dot{x}(t) = f(x(t),z(t)) = g(x(t)) + h(z(t)).
\end{equation}
where the only assumption placed on $h$ is that it is continuous with respect to the time-dependent function $z$. 

The decomposition of the system $f$ into the sum of two different components $g$ and $h$ forms the basis of the KB-Sindy algorithm, which is illustrated in Fig. \ref{FigIntro}.\\

The choice of the variable $z$ depends on the specific problem and can be used to describe different scenarios:
\begin{itemize}
\item a typical case is one in which $z$ is a function of $x$, making $h$ a nonlinear function of state variables. Thus, $g$ and $h$ share the same nature but $h$ is not easily represented as a sum on monomials; 
\item the system may be non-autonomous, subject to measurable inputs $u(t)$ whose effect on the governing equations is unknown. In this case one can set $z=u$ and $h$ describes an unknown transformation of $u$, which must to be estimated from the data;
\item there is feedback of unknown functional form from an observable output $o(x(t))$. In this case, $z(t)=o(t)$ and $h$ describes the unknown feedback channel; 
\item the system is subject to stochastic noise that influences the dynamics, as happens in models of gene regulation where the promoter can exist in different states \cite{Ai2003,AlonUri2020}. We can now think of $z$ as the time variable and $h(t)$ as the unknown realization of the noise at time $t$. 
\end{itemize}

Estimating the functions $g$ and $h$ in
\eqref{SecondStateMod} poses considerable challenges.
For example, the problem of reconstructing $h$ with only knowledge of its regularity from a finite number of measurements is ill-posed. There may also be situations where the number of data points is similar to (or even smaller than) the number of basis functions introduced to describe $g$. This may require the use of additional regularization methods beyond sparsity.
Machine learning techniques based on kernels $\mathcal{K}$ are useful in this situation \cite{Rasmussen,Scholkopf01b}. They implicitly encode a large number of smooth basis functions and control complexity by penalizing estimates of $h$ that are highly irregular and thus physically implausible. These models can also be used to describe colored noise processes whose statistics are unknown, by only including the information that the process is continuous in time, as in the case of Brownian motion. In our setting, $\mathcal{K}$ is a real-valued semidefinite positive symmetric function and, for any pair $z_a,z_b$, the value $\mathcal{K}(z_a,z_b)$ is a measure of similarity between $h(z_a)$ and $h(z_b)$. For example, if $\mathcal{K}$
is continuous, then all functions $h$ compatible with $\mathcal{K}$ will be continuous.  An important example
is the Gaussian kernel, commonly 
used in machine learning to model smooth functions. It is defined by 
\begin{equation}\label{GaussKer}
\mathcal{K}(z_a,z_b) = \rho \exp\Big(-\frac{\|z_a-z_b\|^2}{\omega}\Big), \qquad \rho,\omega > 0
\end{equation}
where the scale factor $\rho$ and the width $\omega$ are hyperparameters
that control the degree of continuity of the estimate for $h$. 
These parameters are typically unknown and must be estimated from the data.
Importantly, the Gaussian kernel is universal: it implicitly embeds an infinite number of functions
which can approximate any continuous map \cite{Micchelli2006}.\\

As discussed in {\bf Methods}, 
despite the nonparametric nature of the model, kernel theory allows not only $g$ but also $h$ to be written as a finite sum of functions. Indeed, using one of machine learning's central results, the representer theorem \cite{Wahba:90}, on
the function 
\begin{equation}\label{SecondModelForf}
f(x,z)= g(x) + h(z)
\end{equation}
one has that
\begin{itemize}
\item the first component $g$ is \emph{parametric} and given by 
$$
g(x)=\sum_{i=1}^p \Xi_i \phi_i(x),
$$
a sum of \emph{explicit} basis functions with complexity controlled by sparsity-promoting methods;
\item the second component $h$ is \emph{nonparametric} and given by
\begin{equation}\label{RT}
h(z) = \sum_{i=1}^m \xi_i \mathcal{K}(z,z(t_i)).
\end{equation}
\end{itemize}
From the above equation, we see that the basis functions for $h(z)$ are the kernel sections centered on $z(t_i)$ where the $t_i$ are the $m$ instants where data are collected. For example, $h(z)$ consist of $m$ Gaussian functions using \eqref{GaussKer}. The nonparametric nature of $h$ is reflected in the fact that the number of unknowns $\xi_i$ is not fixed in advance, being equal to the number of available measurements $m$. The complexity of $h$ is not controlled by the number of active basis functions, but by the smoothness information on $h$ encoded in the choice of $\mathcal{K}$, which also imposes a penalty on the expansion coefficients $\xi_i$ \cite{SurveyKBsysid}. A single function $h(z)$ is considered for simplicity. However, the nonparametric component could be the sum of functions, each associated with its own kernel.

It follows that  \eqref{SecondModelForf} can be expressed in matrix-vector form as follows
\begin{equation}\label{KerMeasModY}
y = \Theta \Xi + K\xi + e.
\end{equation}
Compared to \eqref{MeasModY}, \eqref{KerMeasModY} contains the kernel-induced term $K\xi$ where $K$ is the $m \times m$ kernel matrix with $(i,j)$ entry $\mathcal{K}(z(t_i),z(t_j))$ while
the vector $\xi$ contains the $m$  weights $\xi_i$ which embed the estimate of $h(z)$ for any $z$ through \eqref{RT}.\\
Estimates of $\Xi$ and $\xi$ can be obtained using the KB-Sindy algorithm. It solves a sequence of regularized least squares problems, using Sindy to control the sparsity of $\Xi$
and the kernel to regularize $\xi$. 
The pseudocode, along with strategies for estimating unknown sparsity and kernel parameters (e.g. $\rho$ and $\omega$ in \eqref{GaussKer}) are presented in {\bf Materials and Methods}.

\section*{Results}

\subsection*{Lorenz system subject to time-dependent input and feedback}

The Lorenz model, originally
introduced to describe the unpredictability of weather systems \cite{Lorenz1963}, is a canonical example in chaos theory \cite{Berge1984}.
It has found applications in multiple fields \cite{Strogatz2024}, including ecology, physiology, financial markets, and cryptography \cite{Cushing2003,Freeman1991,Trippi1995,Alvarez2006}.  
The rich dynamics produced by the Lorenz equations and other chaotic systems has led to applications in control theory and signal processing in which these systems are subjected to various forms of input signals and feedback regulation \cite{Ott1990,Chen2003,Lin2005}. The results of these investigations have revealed
new chaotic attractors \cite{Chen1999} and robust adaptive synchronization methods \cite{Yau2000}. 
The Lorenz system was the first example used to benchmark Sindy \cite{Brunton2016}. Here we introduce a more complex version of the identification problem by including a time-dependent forcing term. The system is
\begin{eqnarray}\label{LorenzP2}
\dot{x}_1&=&\sigma(x_2-x_1),\\ \nonumber
\dot{x}_2&=&x_1(\rho-x_3)-x_2 + h(u),\\ \nonumber
\dot{x}_3&=&x_1x_2-\beta x_3,
\end{eqnarray}
where the second equation is subject to a scalar input whose values $u(t)$ are observable. One can also think of $u$ as a manipulable input that can be used to control the system's evolution. However, for such control purposes, it is crucial to understand the effect of $u$ on the system dynamics, which is not assumed known since it is governed by the unknown transformation $h$. Without any information about the system structure, we show that KB-Sindy is able to estimate
the system parameters, preserving the sparsity structure, and recover the input channel $h$.\\ 

\begin{figure*}[h]
	\begin{center}
			{ \includegraphics[scale=0.3]{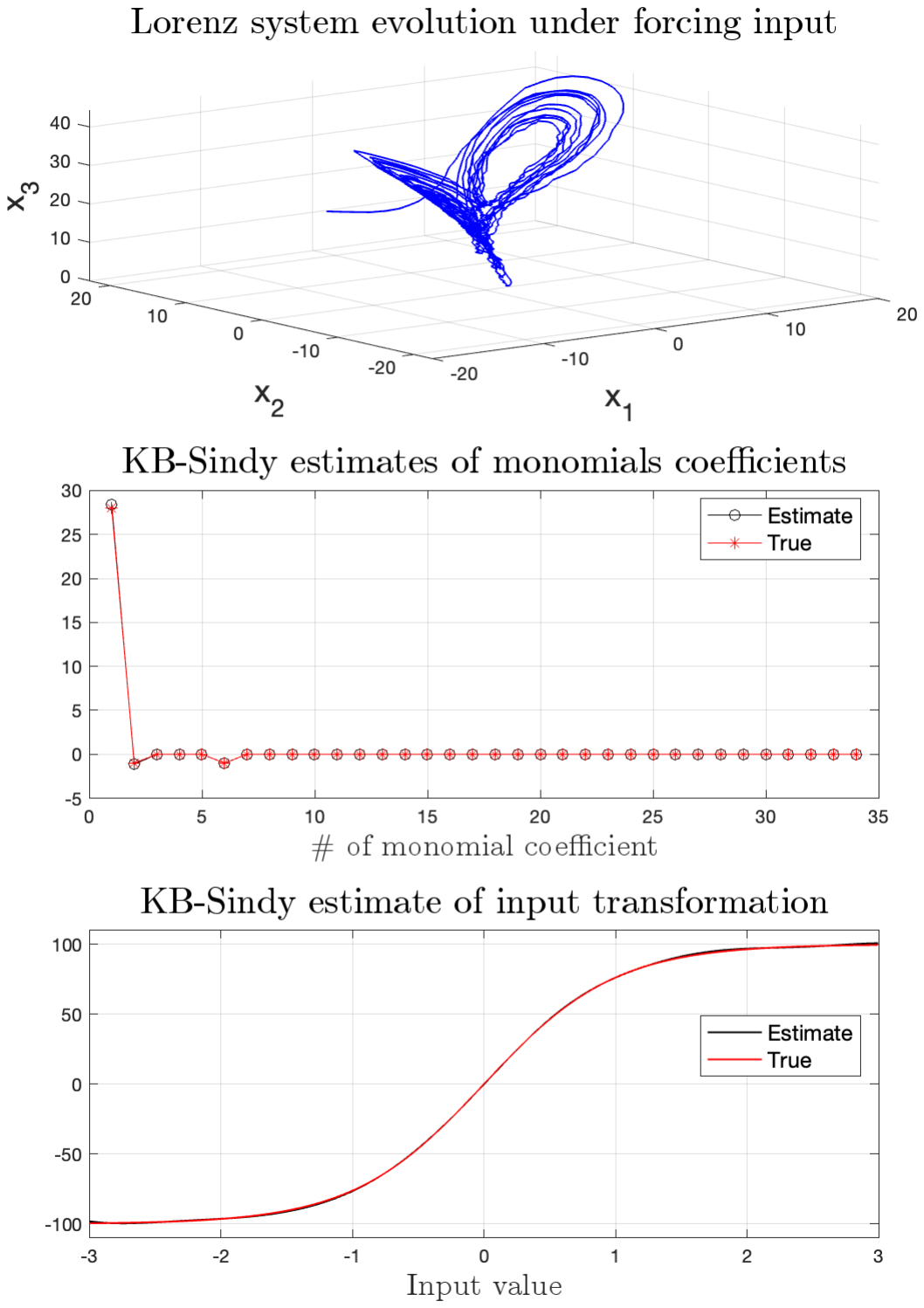}} \ { \includegraphics[scale=0.3]{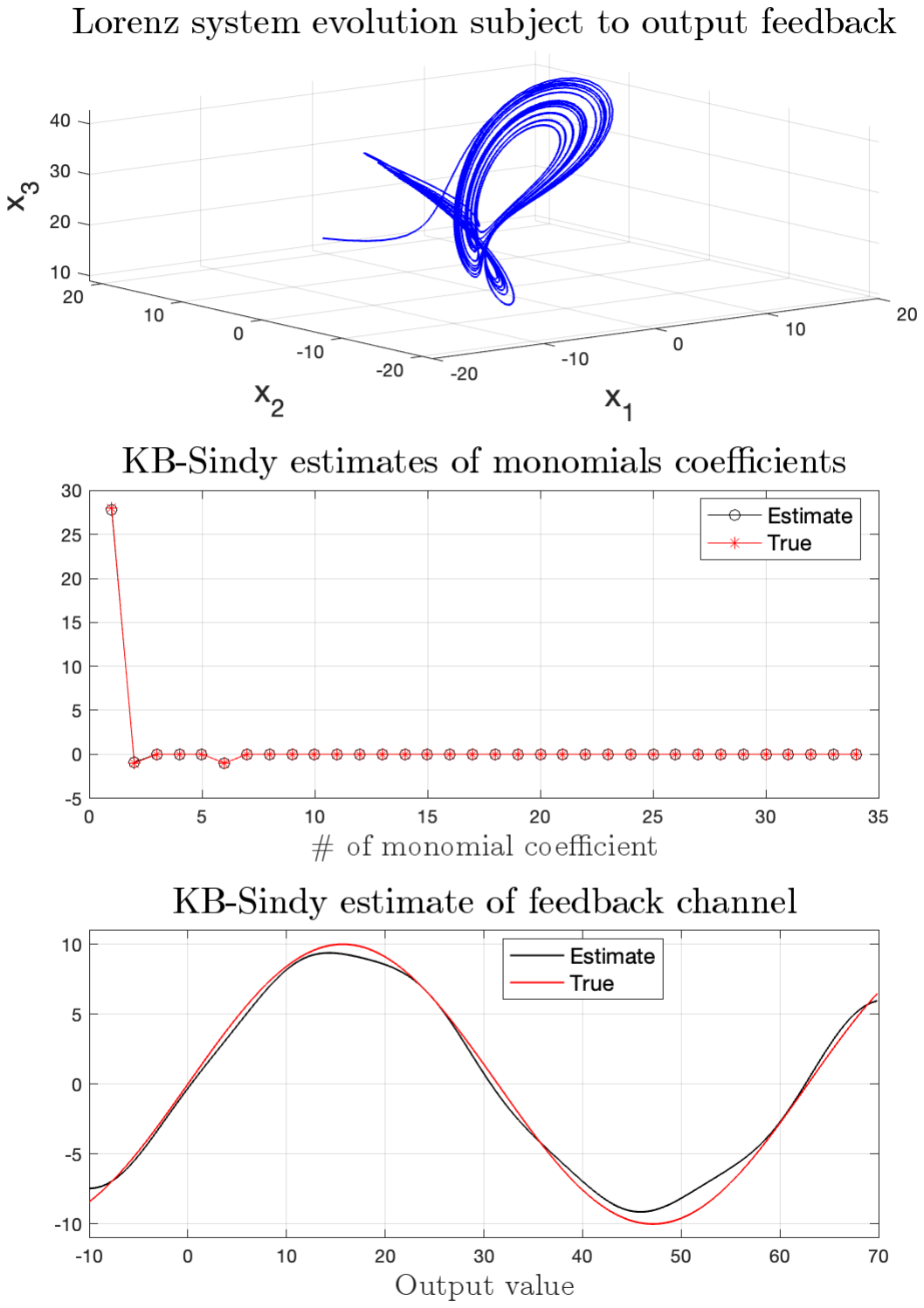}}
		\caption{Lorenz system subject to time-dependent input (left panels) or  output 
        feedback (right). The top panels show the system's time evolution in phase space. 
        KB-Sindy performs joint reconstruction of system parameters and of the nonlinear transformation of the input signal or feedback control. Results are shown for the second equation in \eqref{SecondEqLorenz} with $h$ depending either on the external forcing input $u(t)$ or on the system output $o(t)$ given by the sum of the three states $x_i(t)$. The parametric part of the model contains monomials up to order 4 requiring estimation of 4 coefficients. Only three of them are non-zero and equal to 28,-1,-1. The estimated coefficients are in the middle panels (black while the red circles correspond to the true values).   The monomial coefficients related to $x_1,x_2,x_1x_2$, associated with the coefficients of numbers $1,2,6$, are non-zero  and their estimates are close to the true parameter values.  
        The KB-Sindy accurately captures the nonlinear transformation $h$ for the  input and feedback 
        (bottom panels).}  \label{FigLorenz}
	\end{center}
\end{figure*}

In this example $h$ is proportional to the hyperbolic tangent (Fig. \ref{FigLorenz}, red curve, bottom left panel).
The input $u(t)$ is taken to be constant over time intervals of length $0.1$,
with values generated by independent realizations of a uniformly distributed random variable over the interval $[-3,3]$. The input $u(t)$ can be considered part of an identification experiment, where perturbations are applied to the system to discover the equations that govern the phenomenon being studied.
We focus on estimation of the second Lorenz equation, which
contains only two monomials of order 1 (with coefficients 28 and -1), one of order 2 (with coefficient -1) and the function $h$, i.e.
\begin{equation}\label{SecondEqLorenz}
\dot{x}_2 = 28 x_1-x_2 -x_1 x_3 + h(u).
\end{equation}
The model in \eqref{KerMeasModY} is adopted with the parametric part given by monomials
up to order 4, resulting in a total of 34 unknown coefficients. We then set $z(t)=u(t)$ and use the Gaussian kernel in \eqref{GaussKer} to describe $h(u)$. 
Following \cite{Brunton2016}, we generate 1000 state values simulating \eqref{LorenzP2}   (Fig. \ref{FigLorenz}, top panel) and form a data set with 1000 noisy derivatives. The variance of the noise term $e$ in the estimates for the derivative $y$ defined in \eqref{KerMeasModY},
produces a signal-to-noise ratio (SNR, i.e. the ratio between signal and noise variance) of around $60$.\\ 
Fig. \ref{FigLorenz} shows typical results returned from KB-Sindy for a single
realization of the noise $e$. 
KB-Sindy estimates of the monomial coefficients that describe \eqref{SecondEqLorenz} are close to the true values (Fig. \ref{FigLorenz}, left column, middle row) and the sparsity pattern has been reconstructed. At the same time, KB-Sindy accurately estimates $h$  (Fig. \ref{FigLorenz}, left column, bottom panel). These results demonstrate that KB-Sindy can use sparsity to estimate the part of the system that can be expressed parametrically, while simultaneously estimating the nonparametric
part using only smoothness to reconstruct $h$. 
We also found that this level of performance is preserved even when the SNR is reduced to 9, meaning that the signal is heavily affected by noise. The estimates of the parameters and of $h$ still remain remarkably close to the true values.
We also used KB-Sindy 
to estimate the other two differential equations in \eqref{LorenzP2} using the same settings. That is, including the possibility of other nonlinear transformations $h$ in those equations. KB-Sindy was able to turn off the kernel, learning from the data that no input is present in these equations while providing an accurate estimate of the parametric part of the system. 

We now consider the Lorenz system under output feedback. The
measurable output is $o(t)=x_1(t)+x_2(t)+x_3(t)$ (the sum of the states)
and feeds into the second equation, hence $h(o)$ replaces $h(u)$ in \eqref{SecondEqLorenz}. 
KB-Sindy has to detect the presence of feedback from the data by jointly estimating the system parameters
and the feedback loop $h$ which is taken to be proportional to the sine function (Fig. \ref{FigLorenz}, bottom right panel).
The data are generated as in the previous example, and KB-Sindy is implemented in the same way, now with $z(t)=o(t)$. 
Again KB-Sindy estimates of the monomial coefficients are close to the truth and the sparsity pattern is correctly reconstructed  (Fig. \ref{FigLorenz}, right column, middle panel). At the same time, the functional form of the feedback is well estimated (Fig. \ref{FigLorenz}, right column, bottom panel).
The kernel-based part takes care of the estimation of $h$,
allowing the Lorenz system to be clearly distinguished from the feedback loop.\\

The examples above are for an unknown function $h$ which
depends on a scalar input. The power of the approach becomes even clearer when applied to systems subject to multiple inputs.
This scenario is presented in the section entitled {\bf Additional simulated studies} in {\bf Methods} where a function $h$ is considered that depends on 10 measurable variables.

\subsection*{Autoregulation in gene expression} 
\begin{figure*}[h]
	\begin{center}
			{ \includegraphics[scale=0.45]{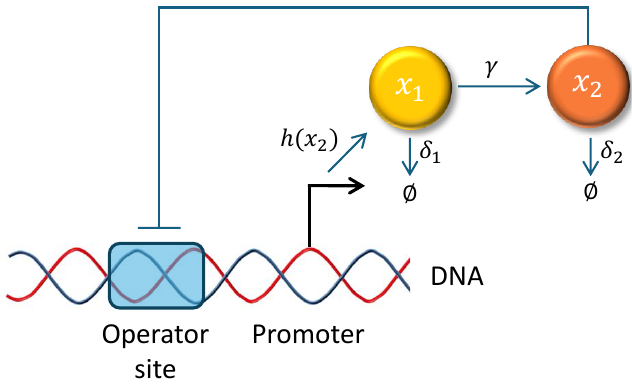}} 
		\caption{Schematic diagram for negative autoregulation. mRNA ($x_1$) is translated into protein ($x_2$) at a rate $\gamma$. The protein acts as a transcriptional repressor by binding  to the gene's promoter. Both  mRNA and protein are degraded at rates $\delta_1$ and $\delta_2$, respectively.}
        \label{FigGeneNetworkNegativeFeedback}
	\end{center}
\end{figure*}

\begin{figure*}[h]
	\begin{center}
			{ \includegraphics[scale=0.3]{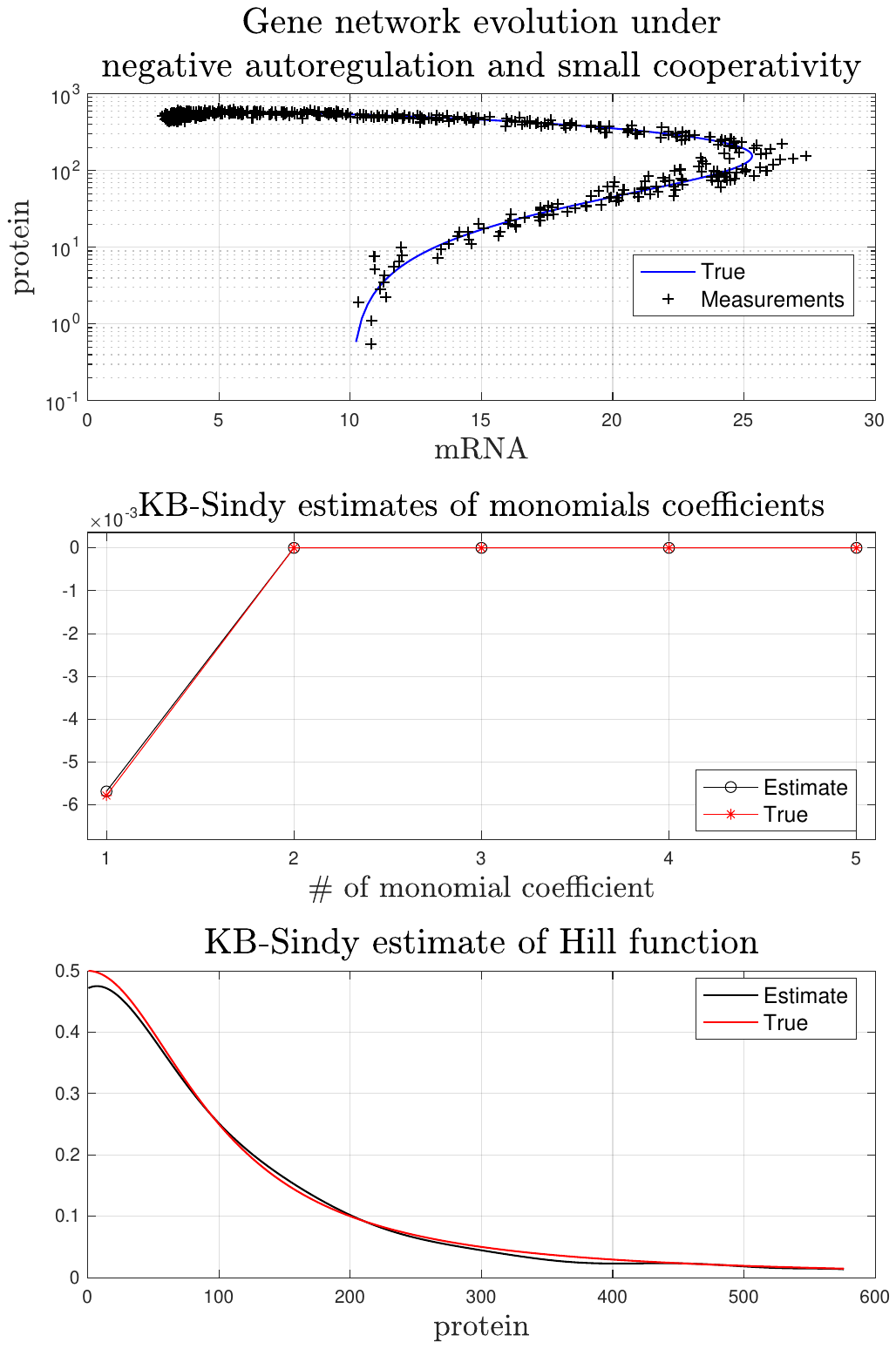}} \ { \includegraphics[scale=0.3]{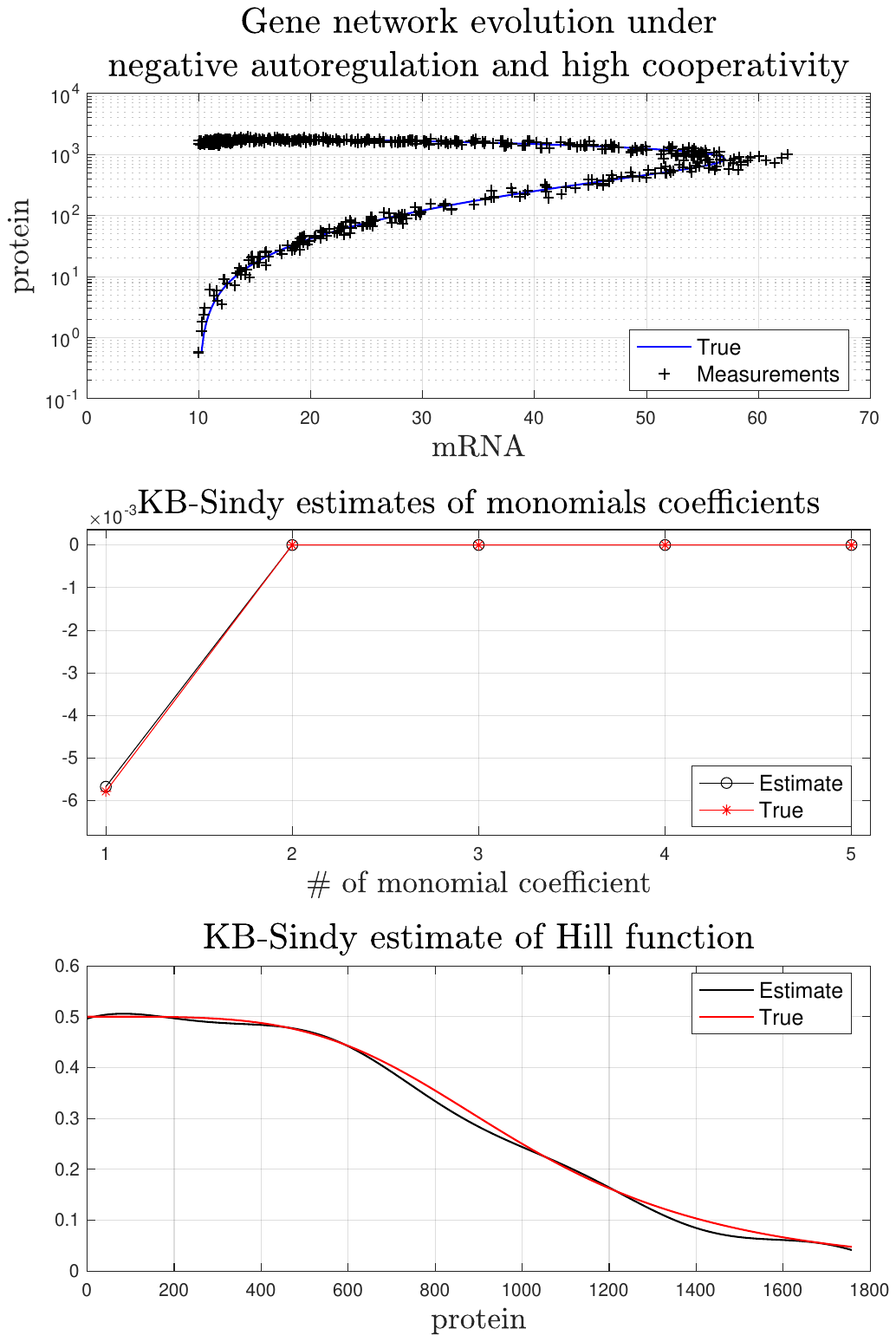}} 
		\caption{Results for negative autoregulation model. The top panels show the simulations of the system using the following parameter values \cite{DelVecchio2014}: $s_1 = 0.5 \, [nM/s]$, $s_2 = 0.5 \cdot 10^{-4} \, [s^{-1}]$, $\delta_1 = 5.78 \cdot 10^{-3} \, [s^{-1}]$, $\delta_2 = 1.16 [s^{-1}] \cdot 10^{-3}$  under weak  ($S = 10^2,q = 2$)  and high ($S = 10^3,q = 4$) nonlinearity. These panels also display the noisy measurements used by 
        KB-Sindy to the parametric model component and the nonlinear Hill function. The estimated coefficients for the parametric component are given in the middle panels (black circles). The red circles correspond to the true values. 
        The estimates for the Hill function $h$ are given in the bottom panels.}  \label{FigHill}
	\end{center}
\end{figure*}

Gene regulatory networks are complex systems in which genes regulate each others expression through transcription factors that bind to the promoter region of the target gene. A common mechanism of gene regulation is autoregulation in which a gene regulates its own expression. 
We focus on negative autoregulation in which the protein encoded by a gene represses its own transcription \cite{AlonUri2020, DelVecchio2014} (Fig. \ref{FigGeneNetworkNegativeFeedback}). Cells use this form of negative feedback for several purposes depending on the time scale associated with the negative feedback. Rapid negative feedback can accelerate that rate at which a cell reestablishes steady state in response to an external stimuli  \cite{Rosenfeld2002} or reduce noise in gene expression \cite{Kaern2005, Singh2011, Thattai2001,ElSamad2021}. Alternatively, delayed negative feedback can generate sustained oscillations in gene expression.  Hill functions provide a good approximation for modeling gene regulation and other types of biochemical reactions.  We present an example that illustrates how the Hill function can be learned from data in a nonparametric way. Consider the following model for negative autoregulation:

\begin{equation}
    \begin{aligned} \label{GeneFB}
        \dot{x}_1 &= - \delta_1x_1+h(x_2), \quad h(x_2)=\frac{s_1}{1 + \left(\frac{x_2}{S}\right)^q},\\
        \dot{x}_2 &= \gamma\, x_1 - \delta_2\, x_2
    \end{aligned}
\end{equation}

\noindent where $x_1$ and $x_2$ are the concentrations of mRNA and protein, respectively. The parameter $s_1$ is the maximum expression rate for mRNA, $\gamma$ is the rate of protein translation,  while $\delta_1$ and $\delta_2$ are the degradation rates for mRNA and protein, respectively. The Hill function $h(x_2)$ models nonlinear negative feedback exerted by the protein on mRNA expression, with the Hill coefficient $q$ determining the steepness of the Hill function and $S$ the protein concentration at which mRNA expression is at its half maximum value \cite{AlonUri2020}.  We consider systems with small and high values of $q$ as shown in  (Fig. \ref{FigHill}, bottom panels, red lines). Our focus is on identification of the equation for mRNA, \eqref{GeneFB},
which contains a first-order monomial, given by $-\delta_1x_1$, and the nonlinearity $h$.
We include monomials up to second order to describe the parametric part and a Gaussian kernel to model $h$ with $z=x_2$. 
After generating time series for the mRNA and protein concentrations by solving \eqref{GeneFB}, both are corrupted by Gaussian measurement noise with a coefficient of variation (CV) equal to
$5\%$. That is, the standard deviation of the noise is $5\%$ of the current value of the state variable. Estimates of the model variables and their derivatives are then obtained by using smoothing splines as detailed in Materials and Methods. 
Fig. \ref{FigHill} shows the results obtained by KB-Sindy using around 600 measurements of $x_1$ and $x_2$ (top panels). The first 100 measurements were collected with a sampling time of 0.5 s, while the remaining ones were collected with a sampling time of 2 s. The results in the middle panel of Fig. \ref{FigHill} for the estimated coefficients for the monomials $\{x_1,x_2,x_1^2,x_1x_2,x_2^2\}$ labeled as 1-5 show that the linear part of the dynamics is well reconstructed. Only the estimate of the first monomial coefficient is different from zero. This means that KB-Sindy learns from the data the absence of nonlinear interaction between $x_1$ and $x_2$ and that Hill function estimation is found entirely from the kernel. Despite the fact that in the two examples the function $h$ behaves quite differently due to different Hill exponents, the regularity information allows KB-Sindy to reconstruct both of them accurately.\\

This example demonstrates the importance of using Sindy and kernels synergistically to extract maximum information from the data. Using only the kernel, as in \cite{Baddoo22}, one could obtain the nonparametric reconstruction of $f$ in \eqref{StateMod}. However, from the estimated surface it would be difficult to interpret the biological processes that drive changes in gene expression.  Alternatively, if the kernel is disabled, Sindy alone does not produce a reasonable result. The reason for this is illustrated by the following Taylor series expansion of the Hill function: 
\begin{equation}
\begin{split}
\frac{1}{(1+x)^4} &= \sum_{n=0}^{+\infty} \frac{1}{6} (-1)^n x^n (n+1)(n+2)(n+3)\\ 
&=1 -4x +10x^2 -20x^3+35x^4 -56x^5 +84x^5
-120 x^6+165x^8 + \text{o}(x^9)
\end{split}
\end{equation}
The above alternating series has increasing coefficients, leading to ill-conditioning. The rapid growth of these terms can overshadow lower degree contributions. This makes 
Sindy numerically unstable. This problem could be circumvented by including more suitable functions in the function library including the  Hill function  \cite{Mangan2016}. However, this requires  a priori information that is not required by KB-Sindy. 
This point is illustrated in the following example  motivated by feedback regulation by human papillomavirus  \cite{Giaretta2019,Giaretta2020}. \\

We consider an example where autoregulation is positive at low protein levels then becomes negative at high protein levels. This behavior is captured by the following function:
\begin{equation}
h(x_2) = \frac{s_2 \, \sigma \, x_2^4 + s_1 \, \xi \, x_2^2 + s_0 \, \lambda}{\sigma \, x_2^4 + \xi \, x_2^2 + \lambda} 
\end{equation}
where $s_0$ represents the basal transcription rate and $s_2$ is the fully repressed transcription rate at high protein concentrations. In between, the function obtains its  maximum value (Fig  \ref{FigHill}, bottom panel). The form of $h(x_2)$, expressed as the ratio of two polynomials, represents an approximation to the master equation for a finite-state Markov model for the promoter of the human papillomavirus under the quasi-equilibrium approximation \cite{Giaretta2020}.\\

Model in \eqref{GeneFB} is simulated with the same parameters as for the case of negative autoregulation but with function $h$ given  above.
Fig. \ref{FigHill} reports the results obtained by KB-Sindy using approximately 600 measurements of $x_1$ and $x_2$. The first 200 measurements were collected with a sampling time of 0.5 s, while the remaining ones were collected with a sampling time of 4 s. After generating the states, they are corrupted by Gaussian measurement noise with a coefficient of variation (CV) equal to $5\%$. The quality of the results is similar to that obtained in the previous case study. Both the parametric part and the complex nonlinearity are accurately reconstructed by KB-Sindy.

\begin{figure*}[h]
	\begin{center}
			{ \includegraphics[scale=0.35]{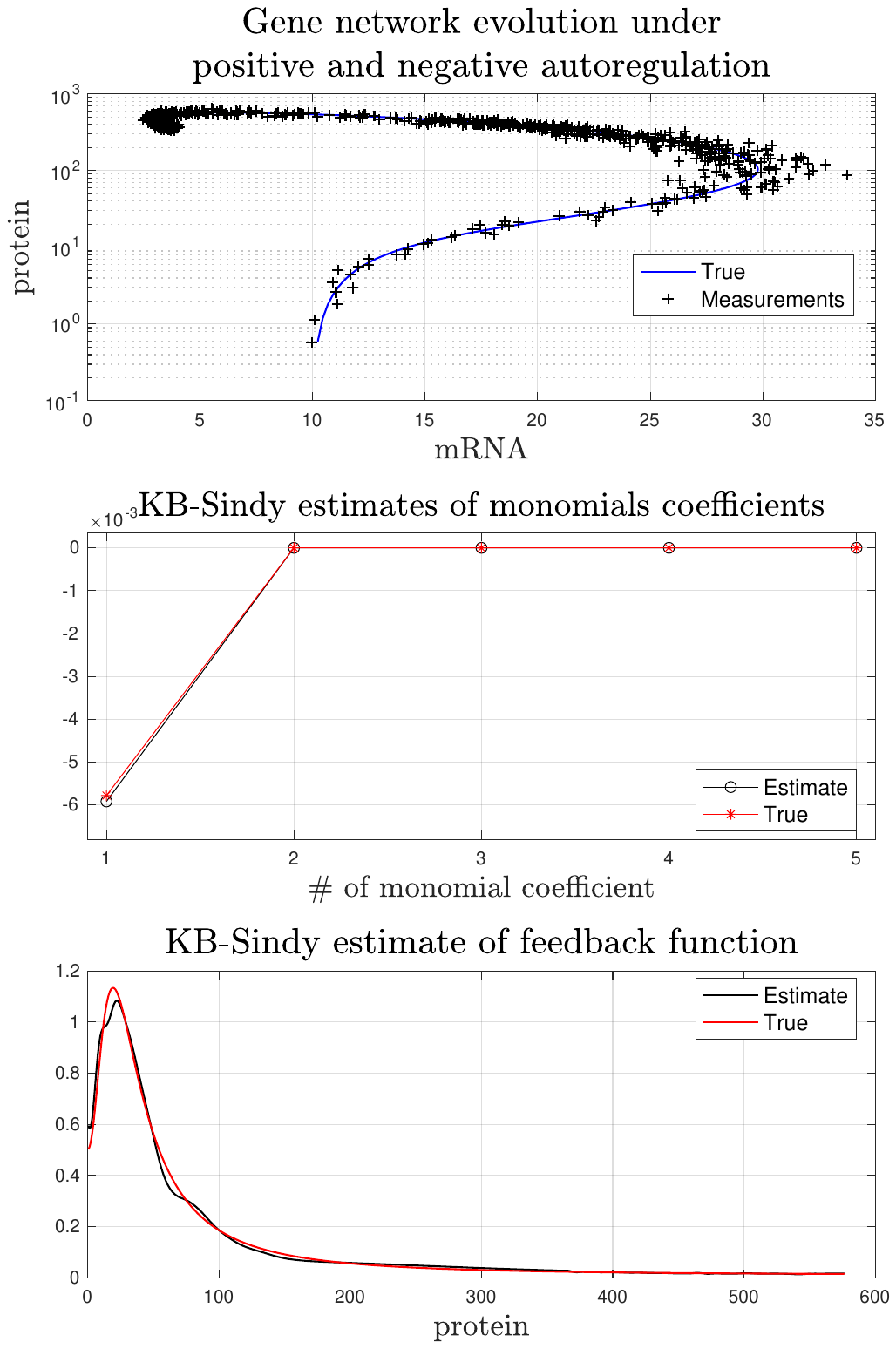}} 
		\caption{Results for autoregulation with positive and negative feedback.  The top panels show the simulations using the same values as the negative autoregulation example   \cite{DelVecchio2014} except with 
        $h(x_2)= (\alpha x_2^4 + \beta x_2^2 + \gamma)/(\sigma x_2^4 + \xi  x_2^2 + \lambda)$ where 
$\alpha=7.36e-11,\beta=1.7e-5,\gamma=0.0011,\sigma=8.9e-9,\xi=8.6e-6,\lambda=0.0022$. 
Also shown are the noisy measurements used by 
        KB-Sindy. The estimated coefficients for the parametric component of the model are given in the middle panels (black circles).  The red circles correspond to the true values. 
        The function $h$ and its nonparametric estimate are shown in the bottom panel.}  \label{FigPosNegFB}
	\end{center}
\end{figure*}

\subsection*{Calcium oscillations} 

Calcium signaling is involved in many cellular processes and exhibits complex spatiotemporal dynamics, such as oscillations and waves \cite{Goldbeter1990, Dupont1991, Dupont1994, Dupont2011, Sneyd1994}.  It is commonly modeled using reaction-diffusion equations (RDEs) that take into account biochemical reactions and diffusion. These models have explained how the local behavior of calcium channels generates global calcium dynamics \cite{Goldbeter1990, Dupont1991}. Because of the prominence of calcium signaling in cell biology, we used it to test KB-Sindy's ability to handle spatial models.\\

We consider a minimal model for 
calcium oscillations based on $Ca^{2+}$-induced $Ca^{2+}$ release \cite{Goldbeter1990, Dupont1991, Sneyd1994}:


\begin{equation}\label{Calcium_real}
\begin{aligned}
\frac{\partial Z}{\partial t} &= \, \, \, 
\begin{tikzpicture}[remember picture, baseline=(diffZ.base)]
\node[anchor=base, inner sep=0pt] (diffZ)
{$D_Z \dfrac{\partial^2 Z}{\partial x^2}$};
\end{tikzpicture}
\, \, \, + \, \, \, 
\begin{tikzpicture}[remember picture, baseline=(nonlinZ.base)]
\node[anchor=base, inner sep=0pt] (nonlinZ)
{$v_0 + v_1 \beta - V_{M2} \dfrac{Z^n}{K_2^n + Z^n} + V_{M3} \dfrac{Y^m}{K_R^m + Y^m} \cdot \dfrac{Z^p}{K_A^p + Z^p} + k_f Y - k Z$};
\end{tikzpicture}, \\
\frac{\partial Y}{\partial t} &= \, \, \, 
\begin{tikzpicture}[remember picture, baseline=(diffY.base)]
\node[anchor=base, inner sep=0pt] (diffY)
{$D_Y \dfrac{\partial^2 Y}{\partial x^2}$};
\end{tikzpicture}
\, \, \, + \, \, \,
\begin{tikzpicture}[remember picture, baseline=(nonlinY.base)]
\node[anchor=base, inner sep=0pt] (nonlinY)
{$V_{M2} \dfrac{Z^n}{K_2^n + Z^n} - V_{M3} \dfrac{Y^m}{K_R^m + Y^m} \cdot \dfrac{Z^p}{K_A^p + Z^p} - k_f Y$};
\end{tikzpicture}
\end{aligned}
\end{equation}

\begin{figure*}
	\begin{center}
			{ \includegraphics[scale=0.55]{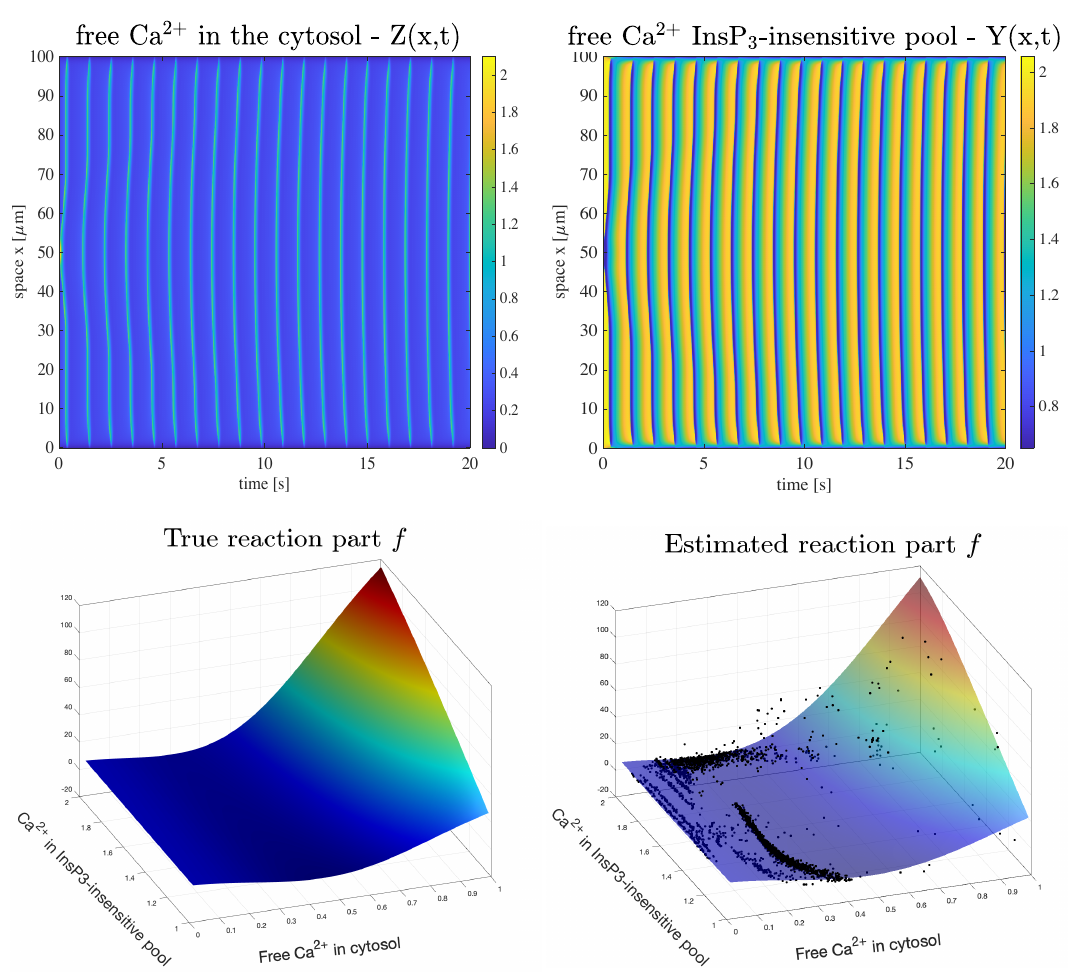}} 
		\caption{Results for calcium dynamics model. Top panels show calcium oscillations generated by \eqref{Calcium_real} using parameter values from \cite{Dupont1991, Sneyd1994}: $v_0=1.0 \mu M s^{-1}$, $v_1=7.3 \mu M s^{-1}$, $V_{M2}= 65 \mu M s^{-1}$, $V_{M3} = 500 \mu M s^{-1}$, $K_r = 2 \mu M$, $K_a = 0.9 \mu M$, $K_2 = 1 \mu M$, $k_f = 1 s^{-1}$, $k = 10 s^{-1}$, $n = m = 2$, $p = 4$, $\beta = 40\%$, $D_Z = 20 \mu m^2 s^{-1}$, $D_Y = 0.1 \mu m^2 s^{-1}$ . The bottom left panel displays the nonlinear function $f$ for the reactions governing cytosolic calcium. The bottom right panel displays  KB-Sindy's estimate for $f$ using the black points shown in this panel.} 
         \label{FigCalcium}
	\end{center}
\end{figure*}

\begin{tikzpicture}[remember picture, overlay]

\node (top_ref) at ([yshift=0.0ex]diffZ.north) {};
\node (bottom_ref) at ([yshift=0.0ex]diffY.south) {};

\node (top_diff) at ([xshift=0pt]diffZ.center |- top_ref) {};
\node (bot_diff) at ([xshift=0pt]diffY.center |- bottom_ref) {};
\node[
  draw=blue,
  fill=blue,
  fill opacity=0.1,
  text opacity=1,
  thick,
  rounded corners=2pt,
  fit=(diffZ)(diffY)(top_diff)(bot_diff),
  inner sep=4pt
] (boxdiff) {};
\path (boxdiff.south) ++(0,-0.3em)
  node[anchor=north, text=blue!80!black, font=\small\itshape]
  {\shortstack{spatial diffusion \\ (parametric)}};

\node (top_nonlin) at ([xshift=0pt]nonlinZ.center |- top_ref) {};
\node (bot_nonlin) at ([xshift=0pt]nonlinY.center |- bottom_ref) {};
\node[
  draw=green!50!black,
  fill=green,
  fill opacity=0.1,
  text opacity=1,
  thick,
  rounded corners=2pt,
  fit=(nonlinZ)(nonlinY)(top_nonlin)(bot_nonlin),
  inner sep=4pt
] (boxnonlin) {};
\path (boxnonlin.south) ++(0,-0.3em)
  node[anchor=north, text=green!50!black, font=\small\itshape]
  {\shortstack{reaction vector fields f(Z,Y) and g(Z,Y)\\ (nonparametric)}};

\end{tikzpicture}\\\\

\noindent where $Z$ and $Y$ describe the concentrations of free $Ca^{2+}$ in the cytosol and an $InsP_3$-insensitive pool located in the endoplasmic reticulum or sarcoplasmatic reticulum.  Both $Z$ and $Y$ are allowed to diffuse with diffusion coefficients $D_Z$ and $D_Y$, respectively.  The parameter 
$v_0$ represents a constant flux of $Ca^{2+}$ from the extracellular environment and $v_1 \, \beta$ models the $InsP_3$-dependent release of $Ca^{2+}$ from a $InsP_3$-sensitive pool. The rates $v_2 = V_{M2} \, \frac{Z^n}{K_2^n + Z^n}$ and $v_3 = V_{M3} \, \frac{Y^m}{K_R^m + Y^m} \cdot \frac{Z^p}{K_A^p + Z^p}$ represent pumping of cytosolic $Ca^{2+}$ into the $InsP_3$-insensitive pool and calcium-dependent release of $Ca^{2+}$ from that the $InsP_3$-insensitive pool into the cytosol, respectively \cite{Dupont1991}. The calcium-induced calcium-release modeled by $v_3$ represents a positive feedback loop that generates oscillations through a substrate-depletion mechanism by rapidly depleting the $InsP_3$-insensitive pool. 
The parameter $k_f$ is the rate constant for the leak of $Y$ back into the cytosol and the parameter $k$ is the rate constant for $Ca^{2+}$ passively moving into the extracellular environment \cite{Dupont1991}.
Using biologically relevant values for the parameters, the model generates calcium oscillations (Fig. \ref{FigCalcium}, top panel). We ran simulations to generate 4 million spatio-temporal grid points for $Z$ and $Y$ using a temporal resolution of $2 ms$ and a spatial resolution of $0.1 \mu m$.\\
In the KB-Sindy 
algorithm we allowed for the possibility of drift in the system. Specifically, 
beyond second-order partial derivatives,
we include in the model equations also first-order partial derivatives with respect to the space of both $Z$ and $Y$. We added noise to the estimates for the spatial derivatives and the time derivative  to produce SNRs of approximately 20 and 10, respectively. We focus on estimating the equation for cytosolic calcium. Inspired by \cite{ScienceAdvances2017}, we set the parametric part to first-order monomials for the spatial derivatives, thus including a total of four basis functions. A Gaussian kernel is then used to model the reaction terms, as they are extremely difficult to capture with a function library. Using 2000 noisy measurements of $\frac{\partial Z}{\partial t}$ distributed over the spatial domain, KB-Sindy infers the equation:
$$
\frac{\partial Z}{\partial t} = 
    \, \, \, 19.11\frac{\partial^2 Z}{\partial x^2}+   \hat{f}(Z, Y)
$$
with $\hat{f}$ displayed in  Fig. \ref{FigCalcium} (bottom, right panel) along with the training data (black dots). KB-Sindy thus learned from the data that only the diffusion coefficient $D_Z$ is non-zero (i.e., there is no drift in the system), providing an estimate close to the true value $D_Z=20$. It also returned an accurate estimate of $f$,
as can be seen by comparing the lower panels of Fig. \ref{FigCalcium}.

\subsection*{Logistic map with stochastic forcing} 
\begin{figure*}[h]
	\begin{center}
			{ \includegraphics[scale=0.3]{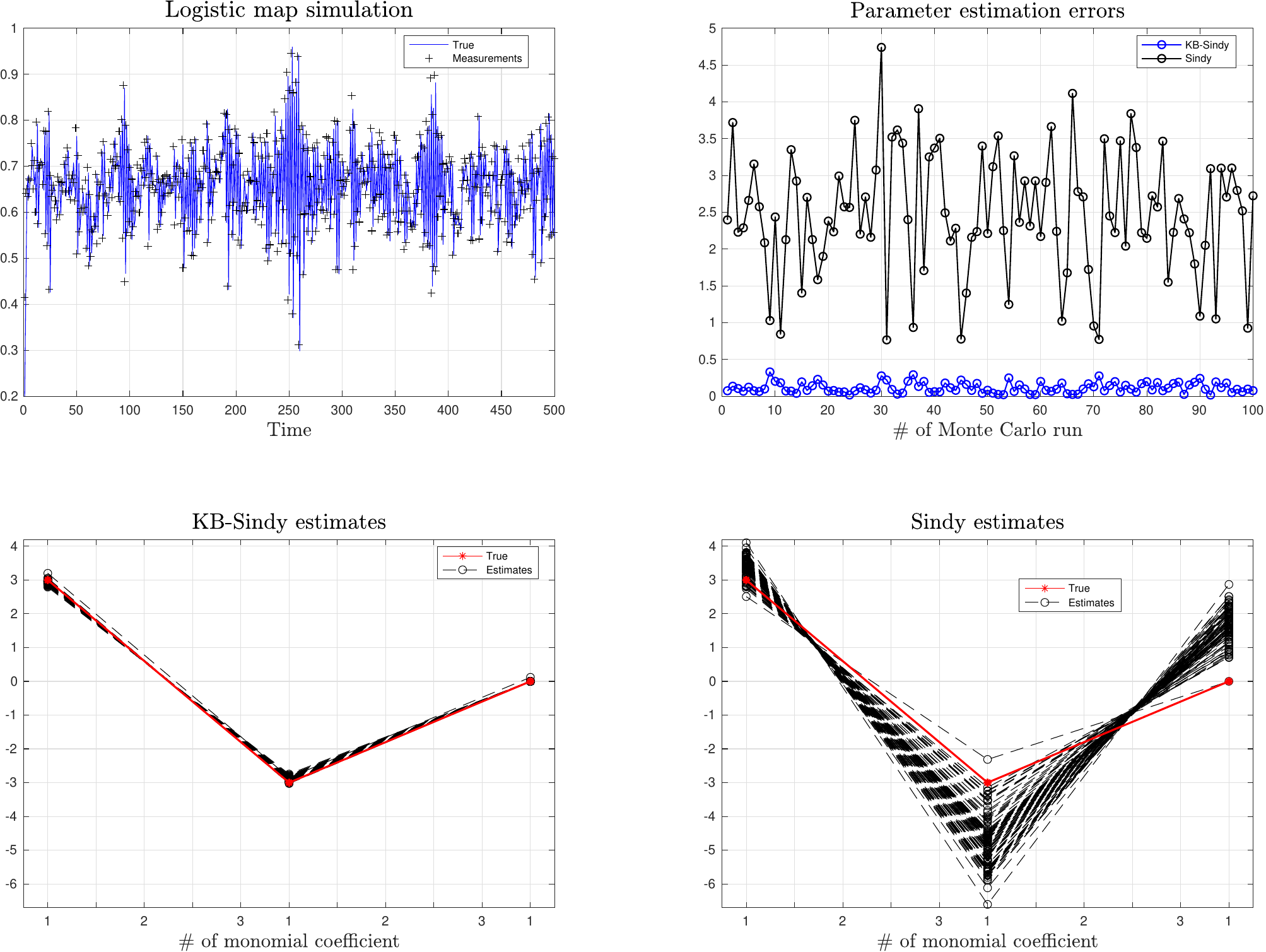}} 
		\caption{Logistic map with stochastic forcing. The top left panel shows a single realization of the system (blue line). The plus signs indicate values after the addition of Guassian noise.  The top right panel shows errors in the estimated parameter for 100 realizations (blue circles - KB-Sindy and black circles - Sindy). Bottom panels show the estimates for the parameters (left panel - KB-Sindy, right panel - Sindy)}  \label{FigLogistic}
	\end{center}
\end{figure*}

The logistic map was popularized in the 1970s by Robert May as a simple discrete-time model for population dynamics that is capable of generating complex behavior \cite{May1976}.  The logistic map is given by $x_{k+1} = r x_k(1-x_k)$, where $r$ is the growth rate. The system exhibits behavior ranging from stable fixed points to chaos \cite{Strogatz2024}.  It has been applied to model insect \cite{Costantino1997} and fish population dynamics \cite{Planque1996}. Stochastic terms can be added to the logistic map to model the affect of environmental fluctuations on the population \cite{Ripa2000}. 
Often capturing the effects of environmental fluctuations requires the use of correlated noise \cite{Halley1996}. Correlated fluctuations have been shown to qualitatively change the behavior of ecological models \cite{Vasseur2004}, \cite{Bjornstad2001} \cite{Ruokolainen2009}. These observations motivate studying the following stochastic system: 
\begin{equation}\label{StochLM}
x_{k+1} = r x_k(1-x_k) + e_k
\end{equation}
where $e_k$ is correlated noise and given by the following autoregressive model:
\begin{equation}\label{StochLM2}
e_{k+1} =0.8 e_k + 0.05 z_k
\end{equation}
with $z_k$ is a Gaussian random variable with unit variance. A simulation including this so called "red noise" \cite{Vasseur2004} and $r=3$ is presented in the top left panel of
Fig. \ref{FigLogistic}. To this time series, we added Gaussian noise with a SNR of approximately 60 (black plus signs). To infer \eqref{StochLM}, KB-Sindy is used with the parametric part including monomials up to third order. The nonparametric part is used to capture the colored noise with $z$ equal to the time variable $k$ and $h$ is estimated using the Gaussian kernel in \eqref{GaussKer}.\\

We simulated a 100 sample paths of the system, each generated using an independent realization of the noise. The top right panel of Fig. \ref{FigLogistic} compares the errors in the estimates for KB-Sindy and Sindy, where the error is defined as the square root of the sum of the squared errors between the true monomial coefficients and the estimates. The bottom panels show plots of the estimates for the 3 parameters for all 100 realizations. Including the kernel allows KB-Sindy to accurately estimate the parameters for all the realizations. In contrast, Sindy produces significant errors for the parameters to compensate for the effects of the colored noise.

We introduced Kernel-based Sindy (KB-Sindy) as a novel model inference framework that extends the original Sindy algorithm by integrating sparse parametric estimation with nonparametric kernel-based methods. This approach overcomes several limitations of the original Sindy algorithm and offers a more flexible and comprehensive method for inferring model equations from data. By combining sparse parametric and nonparametric components, KB-Sindy is able to capture nonlinearities
that are not easily captured by function libraries consisting of monomials. This feature makes KB-Sindy well-suited for application to mathematical models of cellular processes. The examples presented here demonstrate that KB-Sindy can capture a wide range of system behaviors, including complex input transformations, feedback mechanisms, saturation effects, and stochastic dynamics. 
Additionally, the approach mitigates the curse of dimensionality for models with more than a few degrees of freedom by using
kernel-based methods to encode smooth basis functions and  provide regularization for limited data scenarios (see also the section entitled {\bf Additional simulated studies} in {\bf Methods}).\\

In conclusion, KB-Sindy represents a significant advancement in the field of system identification, offering a robust and flexible framework for discovering model equations. By balancing the interpretability of sparse parametric estimation with the flexibility of nonparametric kernel-based techniques, this approach has the potential to enhance our understanding and prediction capabilities across a wide range of scientific and engineering disciplines.

\section*{Methods} 

\subsection{From Sindy to KB-Sindy}

The Sindy pseudocode is reported below. We use $\lambda$ to indicate the sparsity parameter which controls the number of
components of the governing equation to be set to zero at any iteration.\\

{\bf{\large Sindy}}
\begin{algorithmic}[1]
\State $\hat{\Xi}= \big(\Theta^\top \Theta\big)^{-1} \Theta^\top y$ \Comment{Least squares estimate}
\State  Until convergence do 
\State  \indent  smallinds = $|\hat{\Xi}|<\lambda$; \Comment{find small coefficients}
\State  \indent  $\hat{\Xi}(\text{smallinds})=0$ \Comment{and threshold}
\State  \indent  biginds = $|\hat{\Xi}| \geq \lambda$ \Comment{find coefficients significant w.r.t. $\lambda$}
\State  \indent  $\widetilde{\Theta}= \Theta(:,\text{biginds})$ \Comment{regression matrix restricted to significant terms}
\State  \indent  $\hat{\Xi}(\text{biginds}) = \big(\widetilde{\Theta}^\top \widetilde{\Theta} \big)^{-1} \widetilde{\Theta}^\top y$
\Comment{regress dynamics onto remaining terms to find sparse $\Xi$}
\State \Return estimate $\hat{\Xi}$ 
\end{algorithmic}

\medskip

The following pseudocode is for KB-Sindy, which 
solves a sequence of weighted least squares problems. The weights are defined by the matrix 
$A=K + \eta^2 I_{m}$ where $K$ is the kernel matrix induced by $\mathcal{K}$ while $\eta^2$ is the variance of the noise $e$ affecting $y$ in \eqref{KerMeasModY}.
Once the sparse vector $\hat{\Xi}$ has been computed, the coefficients $\hat{\xi}$ characterizing the nonparametric model part 
are obtained via linear regression using $Y- \Theta  \hat{\Xi}$.\\

{\bf{\large KB Sindy}} 
\begin{algorithmic}[1]
\State Define $A=K + \eta^2 I_{m}$ where $K$ is the kernel matrix induced by $\mathcal{K}$  \Comment{Output kernel matrix}
\State $\hat{\Xi}= \left(\Theta^\top A^{-1} \Theta \right)^{-1} \Theta^\top A^{-1} y $ \Comment{Weighted least squares estimate}
\State Until convergence do 
\State \indent smallinds = $|\hat{\Xi}|<\lambda$ \Comment{find small coefficients}
\State \indent $\hat{\Xi}(\text{smallinds})=0$ \Comment{and threshold}
\State \indent biginds = $|\hat{\Xi}| \geq \lambda$ \Comment{find coefficients significant w.r.t. $\lambda$}
\State \indent $\widetilde{\Theta} = \Theta(:,\text{biginds})$ \Comment{regression matrix restricted to significant terms}
\State \indent $\hat{\Xi}(\text{biginds}) = \big(\widetilde{\Theta}^\top A^{-1} \widetilde{\Theta}\big)^{-1} \widetilde{\Theta}^\top A^{-1}y$
\Statex \indent \Comment{regress dynamics onto remaining terms to find sparse $\Xi$}
\State $\hat{\xi} =  A^{-1} \left(y- \Theta  \hat{\Xi}\right)$  \Comment{regress dynamics to find $\xi$ using $\hat{\Xi}$}
\State \Return estimates $\hat{\Xi}$ and $\hat{\xi}$ 
\end{algorithmic}

The algorithm relies on the Sindy strategy to enforce sparsity on the parametric part, coupled with the theory of reproducing kernels \cite{Aronszajn50,Saitoh88} and the representer theorem \cite{Scholkopf01b}. This representer theorem is a central result from machine learning to reduce estimators formulated in infinite dimensional spaces to finite dimensional optimization problems. In our setting, the representer theorem ensures that, for any $z$, the estimate of $h(z)$ is 
$$
\hat{h}(z) = \sum_{i=1}^m \hat{\xi}_i \mathcal{K}(z,z(t_i)).
$$
The interested reader is referred to 
{\bf Methods} for details.\\
\indent Estimates of states and derivatives are obtained by a continuous-time Kalman smoothing filter \cite{Anderson1979}
which implements cubic splines \cite{Wahba:90} whose complexity scales linearly with the number of measurements $m$ \cite{PillonettoInput2006}. For strongly nonstationary signals which exhibit faster variation in the first part of the experiment, the smoothing filter in \cite{Bell2004} has been adopted where the signal
variance is time-varying. It starts from a large value and then smoothly decays to a plateau value. Such a time-course is deescribed by three unknown hyperparameters estimated from the data. The filter also returns the estimation error variance, which defines the scalar $\eta^2$ present in KB-Sindy.\\ 
\indent The output of KB-Sindy depends on the sparsity parameter $\lambda$ and the hyperparameter vector $\theta$, which contains the unknown hyperparameters of the kernel $\mathcal{K}$, e.g. $\rho$ and $\omega$ in the case of the Gaussian kernel described in \eqref{GaussKer}.
Such vectors can be obtained by cross-validation, where the data are divided into a training and a validation set, and
the estimates maximize the prediction of the validation data.
Other criteria that do not require validation data are CP statistics, AIC or BIC, where the concept of degrees of freedom
(dof) is used \cite{Tibshirani2001,Zou2007}. For given values of $\lambda$ and $\theta$, in our setting dof are set to the sum of the trace of the so-called hat matrix given by $KA^{-1}$, which measures the complexity of the nonparametric part, 
and the number of nonzero components of $\hat{\Xi}$, which accounts for the sparse parametric component.
In particular, BIC has been adopted in our experiments, where the unknown parameters solve 
$$
\min_{\lambda,\theta} \ \| y - \hat{y} \|^2 + \hat{\eta}^2 \log(m) \mbox{dof}
$$
with $\hat{y}=\Theta \hat{\Xi} + K\hat{\xi}$ and $\hat{\eta}^2$ to indicate the noise variance estimate returned by the Kalman smoothing filter.

\subsection{Extended description of KB-Sindy}

\begin{figure*}[h]
	\begin{center}
			\hspace{-.3in}
			{ \includegraphics[scale=0.35]{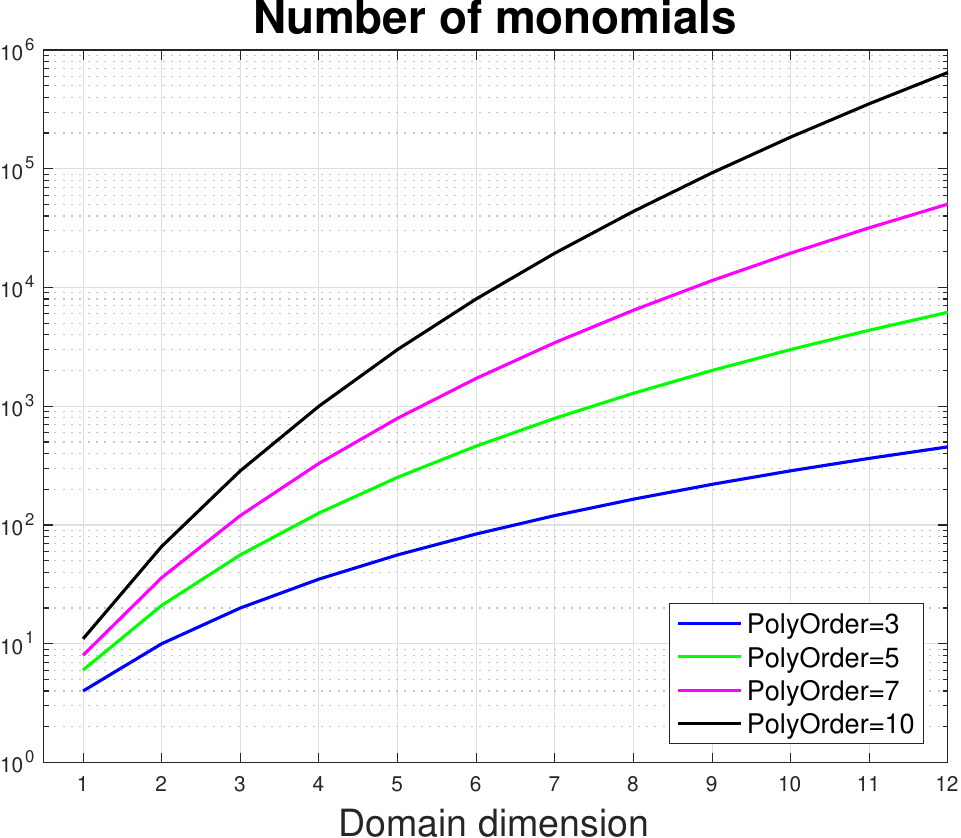}} 
		\caption{Number of monomials contained in a model as a function of state-space dimension or system memory in NFIR models.} \label{FigCoD}
	\end{center}
\end{figure*}

Kernel-based (KB) Sindy integrates
Sindy \cite{Brunton2016} and kernels \cite{Scholkopf01b} in a single algorithm. 
Its development was motivated by the fact 
that 
a possible limitation of Sindy is the need to explicitly introduce many
basis functions to model complex parts of a dynamic system. 
For example, a large number of monomials is needed even for relatively small order $r$.
This is illustrated in Fig. \ref{FigCoD}, which plots
the number of monomials that must be introduced in models of order $3,4$ or $5$.
as a function of the state space dimension or of system memory in models such as nonlinear FIR (NFIR) \cite{Ljung:99}.
A logarithmic scale is used on the $y$ axis.
Due to the curse of dimensionality, it can be seen that the number of monomials grows rapidly 
with model complexity (e.g. measured by state space dimension).
This can lead to situations where the data set size is similar to (or even smaller than) the number of basis functions
introduced into the model, making important the introduction of careful regularisation in the estimation process.
Positive semidefinite kernels are useful in this situation: 
they can implicitly encode a large number of basis functions 
and induce regularisers to control the complexity \cite{Scholkopf01b,SurveyKBsysid,PillonettoPNAS}.
Thanks to this feature, KB-Sindy can also handle situations where an infinite number of basis functions are introduced to describe the governing equations. This versatile estimation framework can  
estimate dynamics 
that are poorly described by a few monomials, 
preserving the sparsity of the solution for other parts of the system
that can be instead well modelled by a small number of basis functions. KB-Sindy can also learn automatically from data 
that complex nonlinearities are not present, thus deactivating the kernel and returning to Sindy.\\
The new identification procedure can also be applied when the system under study is subject to (measurable) inputs 
whose effect on the governing equations is unknown and needs to be estimated from data.
This requires a joint estimation of the system parameters and the input transformation. KB-Sindy does this by 
combining sparse parametric estimation and kernel-based nonparametric estimation by
including only smoothness information on the unknown input channel.
Another important situation concerns systems subject e.g. to output feedback. The problem 
of estimating the unknown feedback loop from data together with the other system parameters has a clear connection
with the previous one, with the input replaced by the system output.

\subsubsection*{Sindy}
An example of the use of the Sindy algorithm 
is the identification of a dynamical system in state space form described by
\begin{equation}\label{StateMod}
\dot{x}(t) = f(x(t))
\end{equation}
where $x(t) \in \mathbb{R}^n$ is the state at time $t$ and $\dot{x}(t)$ contains the derivatives.
Finally, the function $f: \mathbb{R}^n \rightarrow \mathbb{R}^n$ embeds the $n$ unknown state transition maps $f_i$.\\
The main assumptions introduced in \cite{Brunton2016} are that measurements of the state $x$ are available at times $t_1,t_2,\ldots,t_m$
along with a noisy version of the corresponding $\dot{x}$. In addition,  
the system $f$ allows the expansion
\begin{equation}\label{FirstModelFor}
f(x)= \sum_{i=1}^p \Xi_i \phi_i(x)
\end{equation}
by a library of nonlinear basis functions $\phi_i$, reducing
the problem to estimating the unknown expansion coefficients $\Xi_i$.\\
This and many other problems in system identification
can be reformulated in terms of the following linear regression model
\begin{equation}\label{MeasModY}
y = \Theta \Xi + \eta Z
\end{equation}
where $\Theta$ is the known regression matrix, $\Xi$ are the unknown parameters and
$y$ is a column vector containing the available measurements corrupted by white Gaussian noise of SD $\eta$.
In the context of state-space models, the components of $y$ an be seen as the noisy versions of the derivatives 
associated with one of the equations governing the state transition.  
The matrix 
$\Theta$ is often constructed assuming nonlinear system dynamics 
 described by monomials up to a certain order $r$, a model associated with
(truncated) Volterra series \cite{Boyd1985}. 
Sindy can be conveniently used when the vector $\Xi$ containing the unknown monomial coefficients 
is sparse.
%

\subsubsection*{Polynomial and Gaussian kernels}

The basis functions that model 
$f$ in \eqref{FirstModelFor}
and define the regression matrix $\Theta$ in \eqref{MeasModY}
may not be sufficient to describe complex nonlinear systems.
On the other hand, the curse of dimensionality might make it difficult to explicitly include
other model components such as higher-order monomials. 
Furthermore, in some cases it might be desirable to 
introduce a (possibly infinite-dimensional) nonparametric model
which in practice is able to describe all possible nonlinear and smooth dynamics. 
Careful use of regularization is therefore required.\\
The above arguments motivate the study of a more complex situation, where only one part of the physical system is assumed to be sparse and thus describable by a limited number 
of basis functions, while the other one is difficult to model by a parametric structure. As illustrated in the main part of the paper, examples include saturation of state variables, complex feedback mechanisms and transformations of inputs entering the system, stochastic dynamics such as the intrinsic noise described in the systems biology literature. Sparsity takes a back seat in modelling these phenomena, not least because such system components are unlikely to be described by a small number of monomials. The only information available about these functions may be limited to their regularity, such as continuity and differentiability, with respect to a variable $z$ which may represent time, states, inputs and/or outputs. This enriches the original model in \eqref{StateMod} as follows
\begin{equation}\label{SecondStateMod}
\dot{x}(t) = f(x(t),z(t)).
\end{equation}
The transition function $f$ is then modelled as the sum of two parts
which are responsible for capturing different aspects of the phenomenon under study:
\begin{equation}\label{SecondModelForf}
f(x,z)= g(x) + h(z), \quad g(x)=\sum_{i=1}^p \Xi_i \phi_i(x).  
\end{equation}
The first component $g(x)$ is \emph{parametric}, being the sum of \emph{explicit} basis functions with complexity controlled by sparsity-promoting procedures. 
The other, denoted by $h(z)$, is \emph{nonparametric}, associated with a kernel $\mathcal{K}$ which may  \emph{implicitly} contain a very large (possibly infinite) number of basis functions. Its complexity is no
more controlled by the number of active basis functions, but by the smoothness information 
about $h$ encoded in the choice of $\mathcal{K}$. 
We consider a single function $h(z)$ only for simplicity, the nonparametric part could be also sum of functions, each associated with a dedicated kernel.\\ 

\begin{figure*}[h]
	\begin{center}
			\hspace{-.3in}
			{ \includegraphics[scale=0.3]{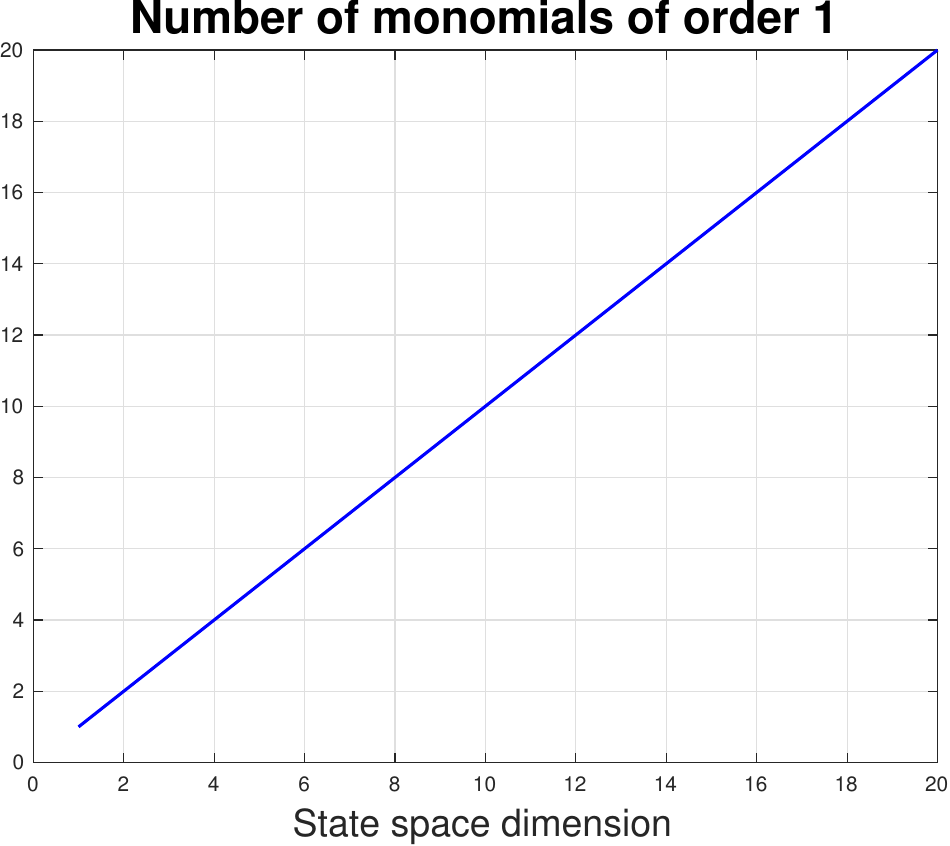}}  \ { \includegraphics[scale=0.3]{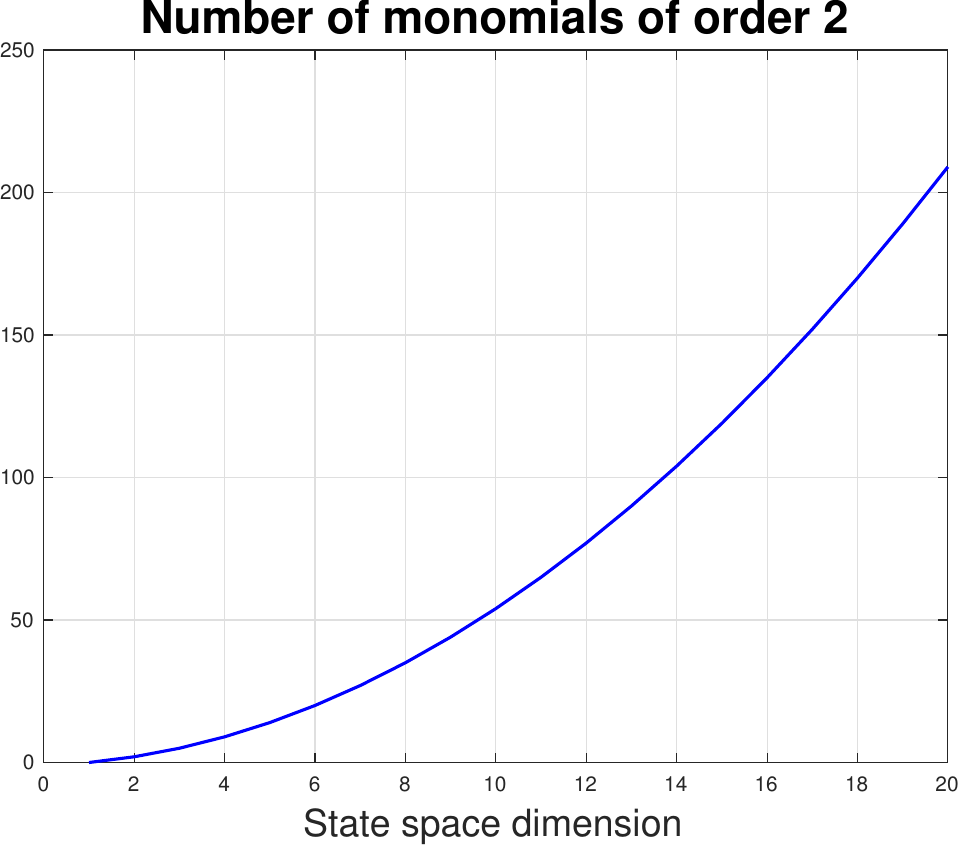}} \\ 
			\ { \includegraphics[scale=0.3]{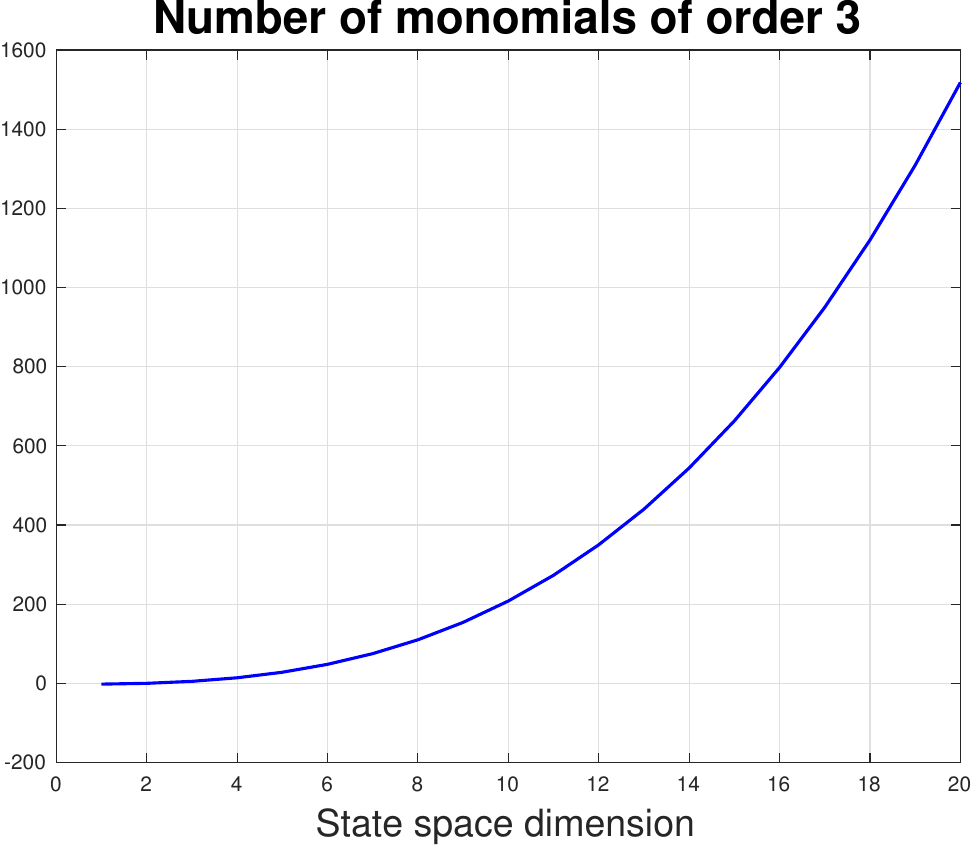}} 
		\caption{Number of monomials embedded in $K_k$, as given in \eqref{dimKkP2}, for monomials order $k$ equal to $1$ (left), $2$ (right) and $3$ (bottom) as a function of state-space dimension} \label{FigCoD234}
	\end{center}
\end{figure*}

As shown in the main part of the paper,
a nonparametric model 
able to describe any reasonable (smooth) dynamics
is defined by the so called \emph{universal kernels} \cite{Micchelli2006}. They  
can approximate arbitrarily well any continuous function and 
an important example
 is the Gaussian kernel. Letting
the integer $m$ denote the dimension of the output vector $y$,
 it induces a $m \times m$ kernel matrix 
$K$ 
 with $i,j$ entry given by
\begin{equation}\label{GaussKer}
K(i,k)= \theta \exp\Big(-\frac{\| z_i-z_k\|^2}{\zeta}\Big)
\end{equation}
where $\zeta$ is an unknown parameter called kernel-width while the
$z_j$ represent $z$ sampled 
at the times $t_1,\ldots,t_m$ where output data in $y$ are collected. For instance,
the $z_j$ could be vectors containing the measured system states or past input values 
when NFIR models are used.\\
Another important kernel that is not universal but may contain
implicitly a very large number of basis functions is 
the polynomial kernel.
When the basis functions $\phi_i$ explicitly introduced   
describe all the
monomials up to order $r$, one can further enrich
the model by introducing homogenous kernels of order larger than $r$
\cite{Franz06aunifying}. 
If the $z_i$ are column vectors of dimension $n$,
each homogenous polynomial kernel of order $j$ induces a kernel matrix 
$K_j$ with $(i,k)$ entry 
\begin{equation}\label{Kkm}
K_j(i,k)= \Big(z_i^{\top}z_k\Big)^j
\end{equation}
which contains implicitly any monomial
of order $j$ sampled at times $t_1,\ldots,t_m$. 
Hence, one has
\begin{equation}\label{dimKkP2}
\# \ \mbox{of monomials ``contained" in} \ K_j=\sum_{k_1+\ldots+k_n=j} \binom{j}{k_1,k_2,\ldots,k_n},
\end{equation}
see also Fig. \ref{FigCoD234} 
which plots \eqref{dimKkP2} 
for $j=1,2,3$ as a function of the state-space dimension. 
We can now introduce the overall kernel matrix
\begin{equation}\label{Kreg3}
K=\theta \sum_{i=r+1}^R K_i
\end{equation}
or  the richer description
\begin{equation}\label{Kreg4}
K=\sum_{i=r+1}^R \theta_i K_i
\end{equation}
where any $K_i$ is given a dedicated nonnegative scale factor $\theta_i$.\\
Assigned a kernel, the model in \eqref{MeasModY} can be now replaced by the 
following more complex description 
\begin{equation}\label{MeasModY2}
Y = \Theta \Xi + K\xi + \eta Z.
\end{equation}
In machine learning the low-dimensional parametric part defined by the regression matrix $ \Theta \Xi$ 
is called the \emph{bias space} and,
in our setting, it is the (explicit) linear combination of monomials up to order $r$.
The other part $K\xi$ embeds (implicitly)  other more complex dynamics, e.g. 
it can capture all the other monomials up to order $R$. 
Thus, \eqref{MeasModY2} describes 
the nonlinear system as sum of two parts,
one in the bias space and the other one in the kernel-induced space.

\subsubsection*{KB-Sindy}

A classical approach to
estimate $\Xi$ and $\xi$ starting from  \eqref{MeasModY2} 
combines
reproducing kernel Hilbert spaces theory \cite{Aronszajn50,Wahba:90} and the extended version of the representer theorem, e.g. see \cite{SpringerRegBook2022}[chapter 6]. In particular,  the representer theorem
ensures that the estimate of the nonparametric part belongs to the subspace 
spanned by the kernel centred on the $z_i$. 
It can be obtained by optimizing an objective where the same matrix $K$ is used as regularizer
for the coefficients $\xi$ whereas the coefficients $\Xi$ of the bias space are not given any penalty.
Specifically, one obtains the following
regularized least-squares problem:
\begin{equation}\label{RegLSext}
({\hat \Xi}, {\hat \xi}) = \arg \min_{\Xi,\xi} \| Y- \Theta \Xi - K \xi \|^2 + \eta^2 \xi^{\top} K \xi
\end{equation} 
where $\eta^2$ is the noise variance and one has
\begin{subequations}\label{ERT}
\begin{align}\label{ERT1}
\hat{\Xi} & = \left(\Theta^\top A^{-1} \Theta \right)^{-1} \Theta^\top A^{-1} Y \\   \label{ERT2}
\hat{\xi} &=  A^{-1} \left(Y- \Theta  \hat{\Xi}\right)\\ \label{ERT3}
A &= K + \eta^2 I_{m}.
\end{align}
\end{subequations}
However, an important issue related to \eqref{ERT} is that sparsity is not promoted when estimating
$\Xi$. If the bias space contains few monomials, this information should be incorporated to improve the estimation process and obtain highly interpretable models. While this could be obtained by adding penalty terms to the objective in
\eqref{RegLSext}, like the $\ell_1$ norm of $\Xi$,
our approach integrates  \eqref{ERT} with Sindy,
leading to KB-Sindy.\\

It is useful to note the following interpretation that underlies the estimation of the bias space
via \eqref{ERT}.  From \eqref{ERT1}, one can see that 
$\hat{\Xi}$ would correspond to (unweighted) least squares
if $A$ were proportional to the identity matrix. In
\eqref{ERT3} one can instead see that $A$ includes not only $\eta^2 I$, the covariance
of the white Gaussian measurement noise of variance $\eta^2$, but also the kernel matrix $K$.
Using the Bayesian  interpetation of regularization \cite{Rasmussen}, one can also interpret $K$ as the covariance of a correlated noise  
describing those nonlinear dynamics which cannot be captured by the explicit basis functions.
This suggests that sparsity can be obtained replacing \eqref{ERT1} with a new version of Sindy which adopts
sequential weighted least squares, with kernel-based weights  defined by the matrix $A$ in \eqref{ERT3}.
This observation leads to the KB-Sindy algorithm 
already summarized in the  pseudo-code reported in the previous section
whose output is  a function of the sparsity parameter $\lambda$ and the hyperparameter vector $\theta$.\\
%
%

One interesting aspect of KB-Sindy is that it boils down to some classical algorithms
for some particular (limit) values of $\lambda$ and $\theta$. In particular,\\

\begin{itemize}
\item if $\theta=0$ the kernel matrix $K$ has all null entries and we recover the original Sindy algorithm. In this case, 
the nonlinear dynamics are described only by the basis functions (e.g. monomials)
explicitly introduced in the model;
\item as $\lambda$ grows to $+\infty$ the estimate of $\hat{\Xi}$ becomes the null vector. Hence,
the bias space is not active and we recover a pure kernel-based approach.
The nonlinear dynamics are expected to lie only in the (high-dimensional) space described implicitly
by the kernel;
\item if $\lambda=0$ no sparsity is induced in the bias space and
we recover the algorithm described in \eqref{ERT}. The estimates of system dynamics are 
the sum of the kernel-based component and of the explicit basis functions with coefficients estimated by least-squares 
(without any sequential procedure). 
\end{itemize}

The vectors $\hat{\Xi}$ and $\hat{\xi}$  permit also to predict the future evolution of the system.
In fact, as already recalled, the representer theorem states that the estimate of the nonparametric part $h$
is the sum of the $m$ kernel sections centred on the $z_i$. It follows that,
for any possible couple $(x,z)$, the estimate of $f(x,z)$ is 
\begin{equation}\label{RepThExt}
\hat{f}(x,z) = \sum_{i=1}^p \phi_i(x)\hat{\Xi}_i + \sum_{k=1}^m 
\mathcal{K}(z_i,z) \hat{\xi}_i.
\end{equation}

When the Gaussian kernel introduced in \eqref{GaussKer} is used, 
\eqref{RepThExt} becomes 
\begin{equation}\label{RepThExtGK}
\hat{f}(x,z)  = \sum_{i=1}^p \phi_i(x)\hat{\Xi}_i + \theta \sum_{k=1}^m \hat{\xi}_k \exp\Big(-\frac{\| z-z_k\|^2}{\zeta}\Big).
\end{equation}
If we instead use  \eqref{Kreg4}, with $K_i$ embedding monomials as defined in \eqref{Kkm}, one then has
\begin{equation}\label{RepThExt2}
\hat{f}(x,z)  = \sum_{i=1}^p \phi_i(x)\hat{\Xi}_i + \sum_{k=1}^m \hat{\xi}_k \sum_{j=r+1}^R \theta_j \Big (z^{\top} z_k \Big)^j. 
\end{equation}
Moving to an even smaller detail level, the estimate of each monomial implicitly embedded in the kernel matrix $K_j$ can be 
also extracted. It can be seen that any monomial $h_{kj}$ of order $j$ entering $K_j$ has a certain weight 
$w_{kj}$ corresponding to the multiplicity with which it appears in the solution. In particular, 
\begin{equation}\label{Weight}
h_{kj}=z_1^{k_1}z_2^{k_2}\ldots z_d^{k_n} \ \Rightarrow \ w_{kj}= \binom{j}{k_1,k_2,\ldots,k_d}.
\end{equation}
Furthermore, the kernel-based component 
$\sum_{k=1}^m \hat{\xi}_k \sum_{j=r+1}^R \theta_j \Big (z^{\top} z_k\Big)^j$ of the 
estimate  in \eqref{RepThExt2} can be decomposed as sum of monomials $\hat{d}_{kj}  h_{kj}(z)$ where 
\begin{equation}\label{RepTh4}
\hat{d}_{kj} = w_{kj} \theta_{k}  H_{kj}^\top \hat{\xi},
\end{equation}
and $H_{kj}$ is the $m$-dimensional (column) vector with components 
defined by the monomial $h_{kj}$ built using $z_i$ for $i=1,\ldots,m$.\\

\subsection*{Additional simulated studies} \label{NumExp2}

The numerical illustrations contained in the main part of the paper
are now complemented with other four experiments.
In some of them, we compare Sindy and KB-Sindy also making use
of a test set to calculate their prediction capability.
In particular, using $y^{test}$ to denote the vector with the derivatives in the test set 
and $\hat{y}^{test}$ the related predictions, the prediction fit is
$$
100\Big(1-\frac{\|y^{test}-\hat{y}^{test}\|}{\|y^{test}\|}\Big).
$$  

\subsubsection{Identification of a Lorenz system subject to ten inputs}

\begin{figure*}[h]
	\begin{center}
			{ \includegraphics[scale=0.35]{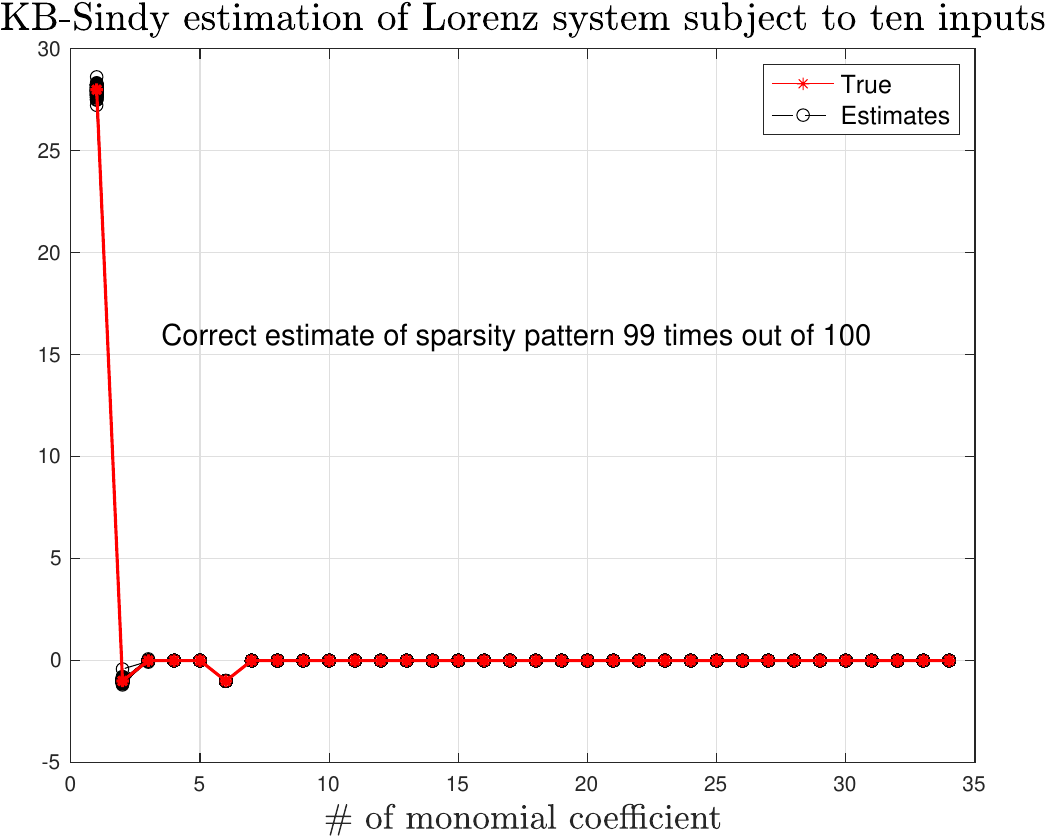}}
		\caption{Lorenz system subject to 10 inputs. The figure reports the 100 estimates of the 34 monomial coefficients which describe the governing equation reported in \eqref{SecondEqLorenzB} (black) from a Monte Carlo study. At any of the 100 runs a new independent noise realization of $e$ is generated
        to form the noisy derivatives contained in the vector $y$ modeled by \eqref{MeasModY2}. The red curve corresponds to the true parameter vector which contains only 3 monomial coefficients different from zero. Despite the presence of various external perturbations given by 10 forcing inputs $u_i$, KB-Sindy understands the nature of these effects and reconstructs the governing equations of a polynomial nature underlying the Lorenz system. In 99 out of the 100 runs the sparsity pattern is perfectly reconstructed with the estimates of the 3 non null monomial coefficients always close to truth.}  \label{Lorenz10inputs}
	\end{center}
\end{figure*}

The examples reported in the main text regarding the identification of a non-autonomous Lorenz system involve an unknown function $h$ which, for the sake of simplicity, 
depends on a scalar input. The power of the approach becomes even clearer when applied to more complex situations with systems subject to multiple inputs $u_i(t)$. For example, consider a function $h$ that depends on 10 measurable variables and influence
the second equation of the Lorenz system as follows:
\begin{equation}\label{SecondEqLorenzB}
\dot{x}_2 = 28 x_1-x_2 -x_1 x_3 + h(u_1,u_2,\ldots,u_{10}).
\end{equation}
Fig. \ref{FigCoD} shows that it is not reasonable to describe $h$ using basis functions, e.g. approximating a nonlinearity of order only 5 requires thousands of monomials. Instead, one can use KB-Sindy with a Gaussian kernel, where now
$z$ contains the 10 inputs:
$$
z(t)=[u_1(t), u_2(t),\ldots u_{10}(t)].
$$
Let $h$ be equal to the hyperbolic tangent whose argument is the mean of the 10 inputs, each assumed measurable and generated by independent realizations from a uniform distribution over $[-3,3]$. The nonparametric part introduced to describe $h$ has no information that the influence from outside depends on the inputs average. Once again it uses the Gaussian kernel which now deals with the curse of dimensionality by processing input locations $z(t)$ of dimension 10. The parametric part of the second Lorenz equation is described by monomials up to order 4. We consider a Monte Carlo study of 100 runs in which at any run different inputs $u$ and noises $e$ are generated. Fig. \ref{Lorenz10inputs} plots the 100 estimates of the 34 monomial coefficients  which can be compared with the red profile corresponding to the true vector. Remarkably, all 100 estimates are close to the true value, and 99 times out of the 100 runs the sparsity pattern is perfectly returned. 

\subsubsection{Identification of four stacked Lorenz systems}

The Lorenz system is described by the following
set of nonlinear dynamical equations
\begin{eqnarray}\label{LorenzP2}
\dot{x}_1&=&\sigma(x_2-x_1),\\ \nonumber
\dot{x}_2&=&x_1(\rho-x_3)-x_2,\\ \nonumber
\dot{x}_3&=&x_1x_2-\beta x_3.
\end{eqnarray}
Now consider the identification of a challenging nonlinear system of dimension 12
which is obtained by stacking four Lorenz systems, each described in \eqref{LorenzP2}.
The four systems share the same parameters taken from 
 \cite{Brunton2016}, also used in the main part of the paper, 
but different (randomly chosen) initial conditions.
For each of the 12 state transitions 
we generate a data set of 2000 noisy derivatives 
and associated state values. It is divided into a training and a validation set, 
with 1000 data each. 
The noise SD $\eta$ in \eqref{MeasModY} 
defines a signal-to-noise ratio (SNR) around $60$. We also generate a test set of
2000 noiseless derivatives 
using a different initial condition
to make prediction more difficult. The four Lorenz systems are decoupled, but the identification process does not exploit this information. 
We concentrate on the estimation of the second Lorenz equation, which
contains only two monomials  
of order 1 (with coefficients 28 and -1) and one of order 2 (with coefficient -1).\\

Let us model the system by monomials up to order 3. 
This introduces 442 basis functions with only 1000  
training data. In this case, the classical Sindy algorithm cannot give reasonable results.
 A Monte Carlo study shows that in most cases it returns null estimates
of all monomial coefficients. We can then use KB-Sindy
with a bias space that contains only a reasonable number of basis functions,
allowing the kernels to capture (if necessary) the other components. For this purpose 
grids are used to optimize the hyperparameters. The sparsity parameter
$\lambda$ can vary over $\verb"logspace(-2,1,10)"$, while the components of $\theta$
over the set $[0 \ 0.001 \ 0.01 \ 0.1 \ 1 \ 10]$.\\

\begin{figure*}[h]
	\begin{center}
			{ \includegraphics[scale=0.3]{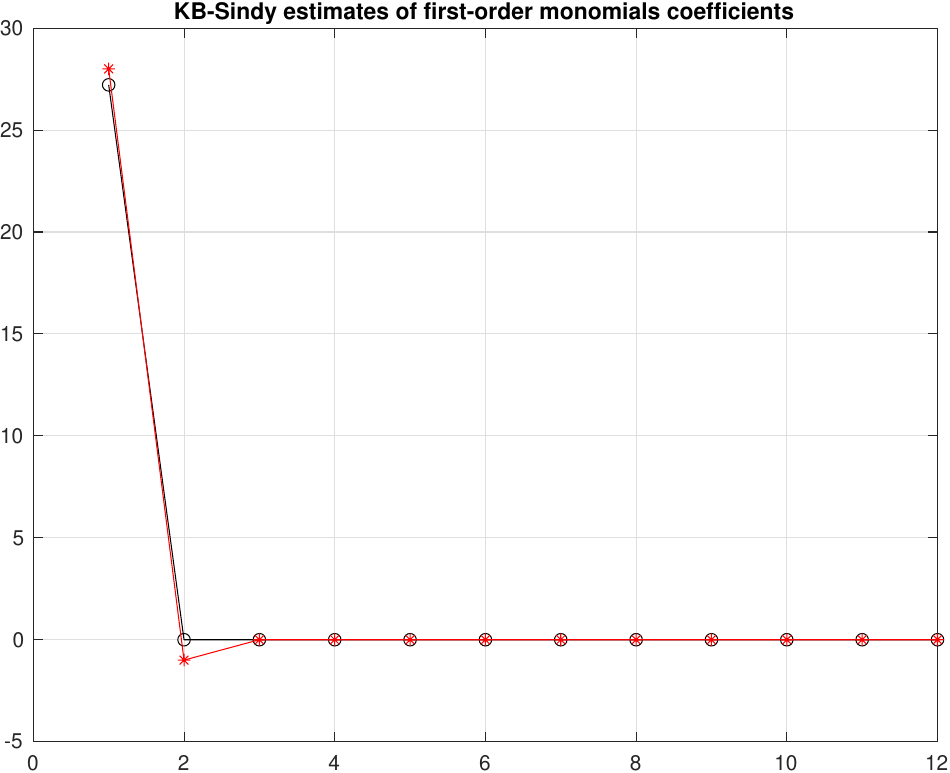}}    \ { \includegraphics[scale=0.3]{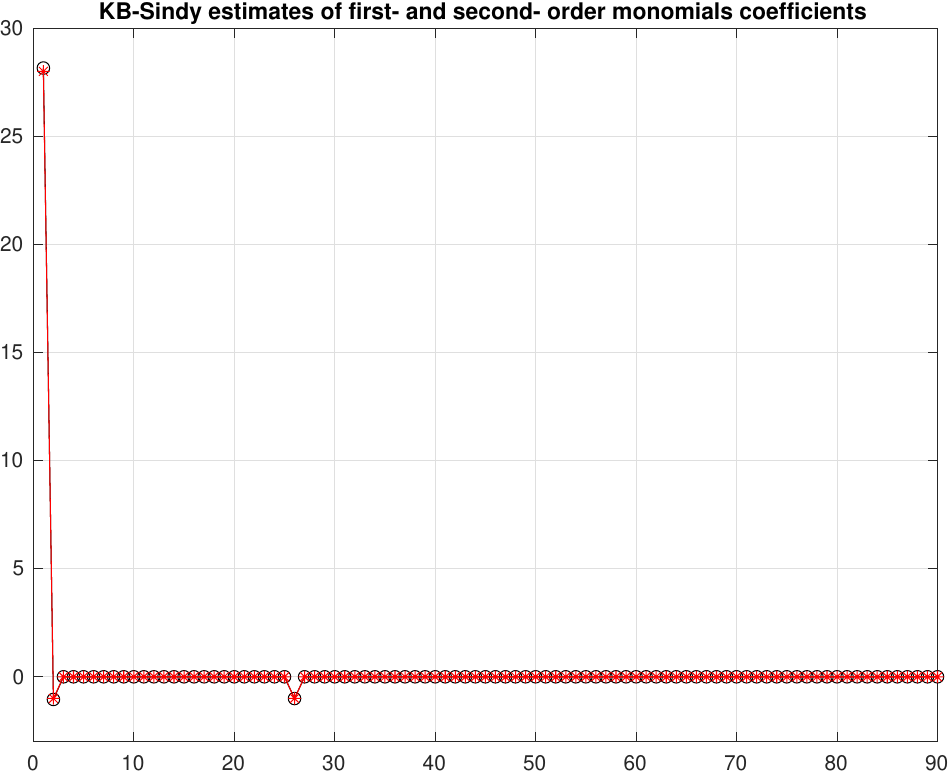}} \\
		\caption{Estimation of the second equation of a system composed of four stacked Lorenz systems. \emph{Left} Estimates of first-order monomial coefficients returned by Kernel-based Sindy using a bias space with monomials up to order 1 and kernels of order 2 and 3. 
		The red circles correspond to the true values. \emph{Right} Estimates of first- and second-order monomial coefficients returned by Kernel-based Sindy using a bias space with monomials up to order 2 and kernels of order 3 and 4. The estimates are so obtained starting from a model which contains 1717 monomials. The same estimates are obtained complementing the bias space with a universal (Gaussian) kernel.}  \label{FigKBSa}
	\end{center}
\end{figure*}

\paragraph{Bias space of order 1} First, we assume a polynomial model of order $r=1$ for the bias space, and then we add two
kernels describing monomials up to order $R=3$ using \eqref{Kreg4}. 
Thus, only 12 explicit basis functions are used,
while the model embeds 442 basis functions in total.
KB-Sindy returns the following hyperparameter estimates: $\hat{\lambda}=2.1,\hat{\theta}_1=0.01,\hat{\theta}_2=0$.
The procedure has correctly recognised that there are only monomials up to second order in the system.
The prediction fit is about $90\%$, showing that KB-Sindy makes good use of the kernels to compensate for what is missing in the 12 explicitly included basis functions.
The estimated coefficients are shown in the top panel of Fig. \ref{FigKBSa}. KB-Sindy promotes sparsity and provides an accurate estimate of the main coefficient, which has a value of 28. The second,
which is equal to -1, is instead incorrectly set to zero.\\

\paragraph{Bias space of order 2 and kernels up to order 4} Now we take a polynomial model of order $r=2$ for the bias space and then add two
kernels which now describe monomials up to order $R=4$ using \eqref{Kreg4}. In this case, 90 basis functions  
are explicitly included and in total the model embeds 1729 basis functions.
KB-Sindy returns the following hyperparameter estimates: $\hat{\lambda}=0.04,\hat{\theta}_1=\hat{\theta}_2=0$.
The algorithm has again correctly learnt from the data that only monomials up to second order are part of the system, but
unlike the previous setting, they all belong to the bias space.
Model complexity control allows
KB-Sindy to be reduced to Sindy which has now 
to handle a restricted number of (explicit) basis functions.
This allows sparsity over all monomials, further improving the estimation process.
The prediction fit is $99.4\%$ and the coefficient estimates are all close to the truth,
see the bottom panel of Fig. \ref{FigKBSa}.\\

\paragraph{Bias space of order 2 and Gaussian kernel} But \emph{can we learn from the data if the Lorenz system contains monomials of even higher order or non-polynomial parts?} We can use KB-Sindy to answer this question, again using a bias space of order $r=2$,
now supplemented with the Gaussian kernel \eqref{GaussKer}.
In a sense, this is equivalent to setting $R$ to $\infty$. The cross-validated score is now optimized 
to estimate the two scalars $\lambda,\theta$ (over the same grids as above) and the kernel width $\zeta$
(over the same grids as above, just without the null value). 
The estimates of $\lambda,\theta$ are the same as before, i.e. $\hat{\lambda}=0.04,\hat{\theta}_1=0$. 
So KB-Sindy suggests 
that there are no other system components outside the bias space.

\subsubsection{Identification of a nonlinear FIR}

Consider the following nonlinear FIR system which is a more complex version of a classical benchmark problem taken from 
\cite{Pillonetto:11nonlin}:
\begin{eqnarray}\label{NfirP2}
y_t&=& \alpha(u_{t-1}+0.6u_{t-2}+0.35u_{t-3}+0.9u_{t-4}+0.35u_{t-5}+0.2u_{t-6}+0.2u_{t-7}  \\ \nonumber
     &+&       u_{t-1}^2+0.25u_{t-4}^2+0.25u_{t-1}u_{t-2}+0.5u_{t-1}u_{t-3}-u_{t-2}u_{t-3}\\ \nonumber
            &+& 0.5u_{t-2}u_{t-4} + 0.1u_{t-3}^3\big) + 0.2 u_{t-1}^4 +\eta e_t
\end{eqnarray}
where $e_t$ and $u_t$ are (independent) white Gaussian noises of unit variance while  $\eta=0.5$.\\
The outputs $y_t$ depend on the past inputs $u_{t-1},\ldots,u_{t-7}$.
So the system memory is 7, a value 
which plays an analogous role to the state space dimension.
Assigned a value of $\alpha$, as specified later, 
a training and validation set of 1000 output data each 
and a test set of size 2000 are generated, starting from null initial conditions.\\ 
In real-world applications, system memory 
is typically unknown. So one might be forced to set it to a conservative (large) value,
then controlling the complexity via hyperparameter estimation. In this example the system memory
is set to 10.  If we use monomials up to order 5,
it follows that there are 3002 basis functions in the model. This 
makes the implementation of Sindy difficult and leads to another interesting case
to evaluate KB-Sindy's ability to control complexity.\\
We consider two situations: $\alpha=1$, which results in a model containing a significant number of monomials distributed over all orders $\{1,2,3,4\}$, and the case 
$\alpha=0$, where the model is really sparse and reduces 
to the single monomial of order 4 given by $0.2 u_{t-1}^4$. KB-Sindy is equipped with a bias space 
with monomials up to order $2$ and with three monomial kernels of order 3, 4 and 5 
associated with three different scale factors $\theta_i$. Three different versions of Sindy are also tested
with monomials of order up to $2,3$ or $4$.\\
For hyperparameter estimation, gradient descent cannot be used in general, because the Sindy nature  
makes the cross-validated objective a non-smooth function of the sparsity parameter $\lambda$. 
The suboptimal solution adopted here is to first set $\lambda=0$, thus disabling the
sparse promotion on the bias space, and then optimizing $\theta$ using gradient descent.
Finally, we set $\theta$ to its estimate and optimise $\lambda$,
thus reactivating the sparse regularisation of the bias space.
Overall, hyperparameter estimation is decomposed into two subproblems.
The first involves smooth optimization to recover the kernel scale factors, the 
second only optimizes the sparsity parameter $\lambda$ over a one-dimensional grid.\\

\begin{figure*}[h]
	\begin{center}
			{ \includegraphics[scale=0.35]{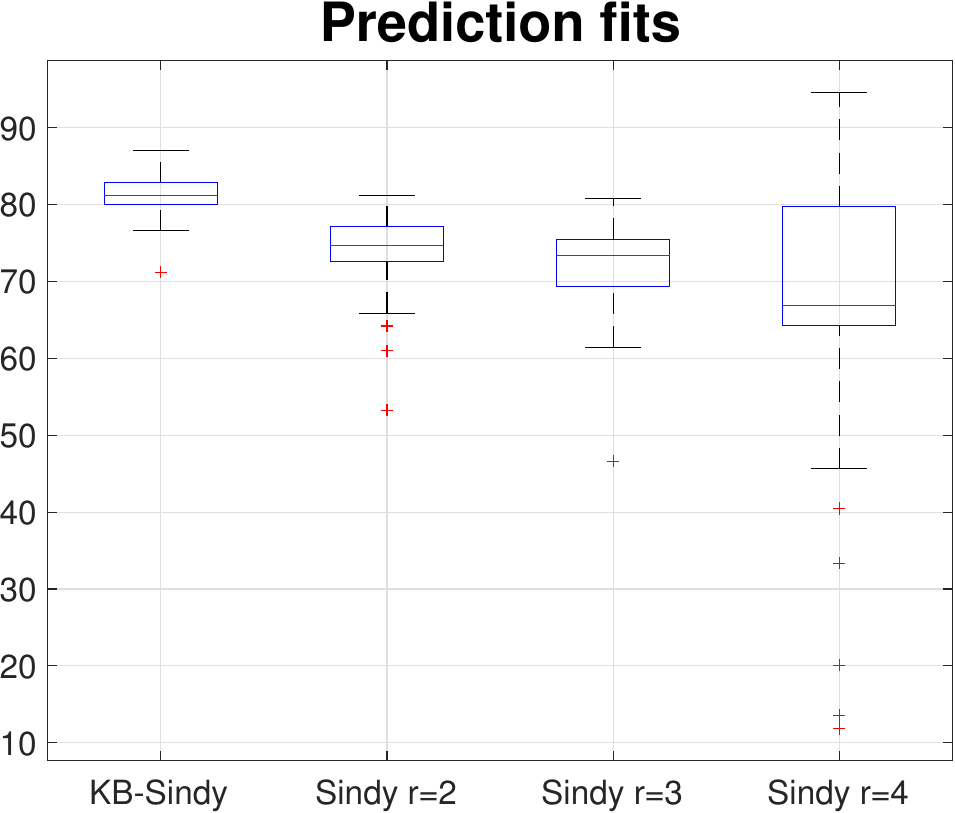}}  
		\caption{Estimation of the NFIR \eqref{NfirP3}. Boxplot of 100 prediction fits from KB-Sindy with a bias space of order $r=2$ and kernels up to order $R=5$ and from Sindy with monomials up to order $2,3,4$.}  
		\label{NFIR1}
	\end{center}
\end{figure*}

\paragraph{The case $\alpha=1$} The system becomes
\begin{eqnarray}\label{NfirP3}
y_t&=& u_{t-1}+0.6u_{t-2}+0.35u_{t-3}+0.9u_{t-4}+0.35u_{t-5}+0.2u_{t-6}+0.2u_{t-7}  \\ \nonumber
     &+&       u_{t-1}^2+0.25u_{t-4}^2+0.25u_{t-1}u_{t-2}+0.5u_{t-1}u_{t-3}-u_{t-2}u_{t-3}\\ \nonumber
            &+& 0.5u_{t-2}u_{t-4} + 0.1u_{t-3}^3 + 0.2 u_{t-1}^4 +\eta e_t.
\end{eqnarray}
We carry out a Monte Carlo study calculating, for each run, the prediction fit 
obtained by the 4 adopted estimators. 
Fig. \ref{NFIR1} shows the boxplot of the 100  
fits. It can be seen that the average prediction performance of KB-Sindy is better than that of
Sindy for all possible implementations.\\

\begin{figure*}[h]
	\begin{center}
			{ \includegraphics[scale=0.35]{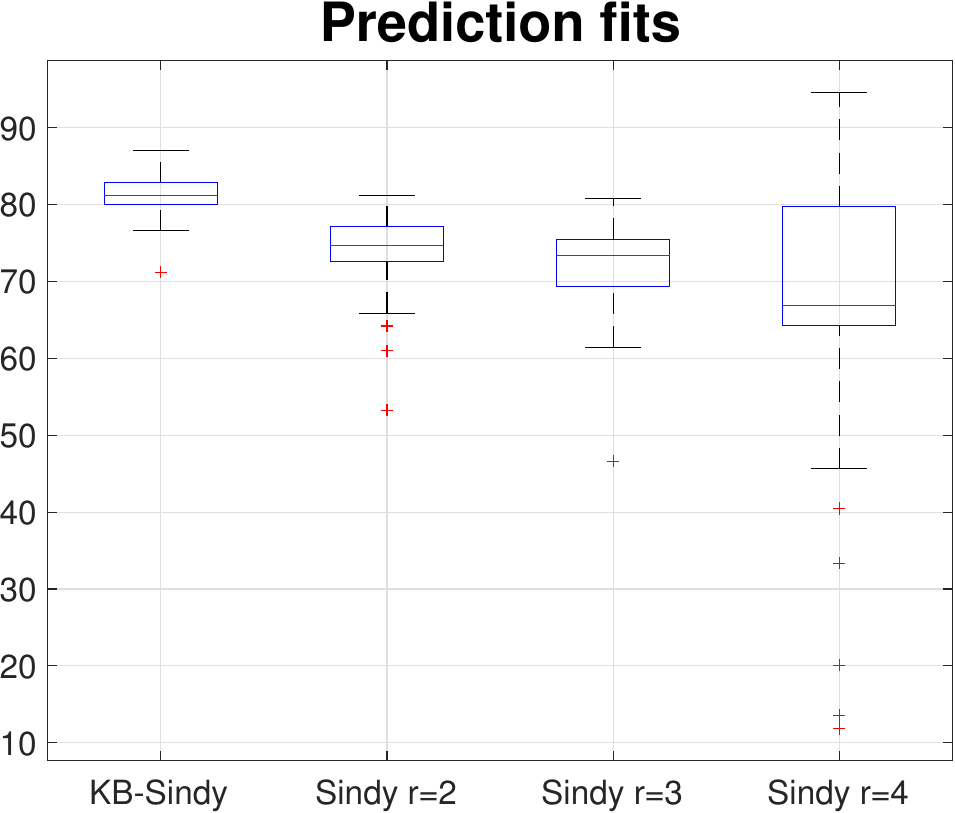}}  \   { \includegraphics[scale=0.35]{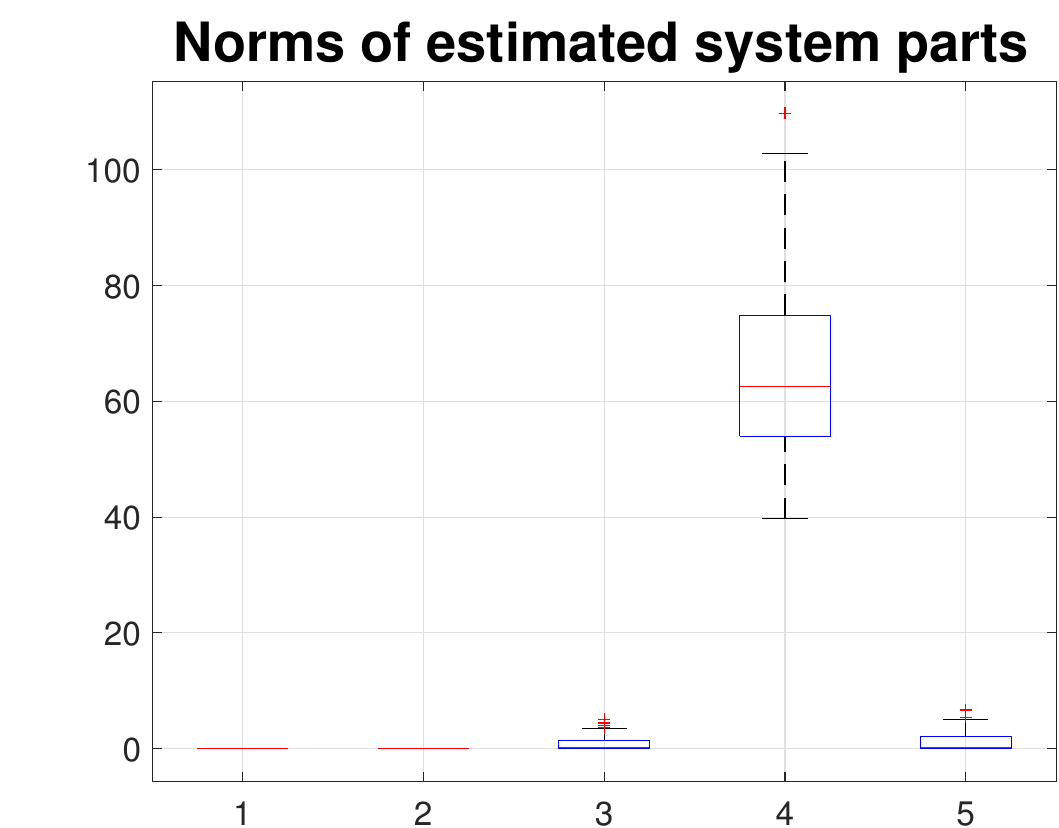}} 
		\caption{Estimation of the NFIR \eqref{NfirP4}. \emph{Left} Boxplot of 100 prediction fits from KB-Sindy with a bias space of order $r=2$ and kernels up to order $R=5$ and from Sindy with monomials up to order $2,3,4$. \emph{Right} Norms of the 100 system parts returned by KB-Sindy as a function of the monomials order.}  
		\label{NFIR2}
	\end{center}
\end{figure*}

\begin{figure*}[h]
	\begin{center}
			{ \includegraphics[scale=0.35]{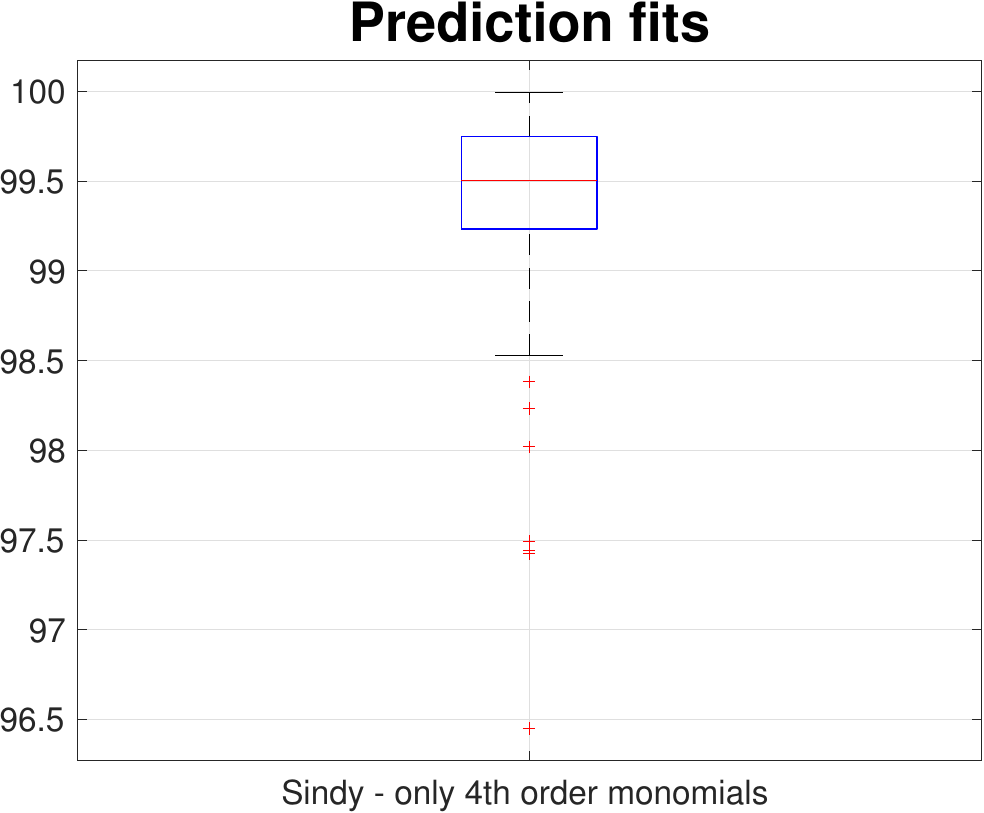}}  
		\caption{Estimation of the NFIR \eqref{NfirP4}. Prediction fits of Sindy using only monomials of order 4, as suggested by the sparsity pattern returned by KB-Sindy. }  
		\label{NFIR3}
	\end{center}
\end{figure*}

\paragraph{The case $\alpha=0$} The system now reduces to
\begin{equation}\label{NfirP4}
y_t= 0.2 u_{t-1}^4 +\eta e_t.
\end{equation}
Again, we perform a Monte Carlo study, calculating, for each run, the prediction fit 
returned by the 4 estimators. 
Fig. \ref{NFIR2} (left panel) shows the boxplot of the 100  
fits. As in the previous case, we can see that (on average) the prediction performance of KB-Sindy is better 
than any of the other three Sindy implementations.\\ 
For this case study, we now show that the results of KB-Sindy can be used by Sindy
to further improve its prediction performance. Remarkably, the synergistic use of these two methods
allows to increase the average 
prediction accuracy to $99\%$.
In order to explore the degree of sparsity of the system under study, at any run 
the estimate of the dynamical system $f$ is decomposed into the sum
$f_1+f_2+\ldots+f_5$ where each $f_i$ contains only monomials of order
 $i$. 
 Next, each $f_i$ is evaluated using the state values
contained in the training set. 
The right panel of the same figure 
shows the boxplots of the Euclidean norms computed after the 100 runs as a function of the monomials order. One can see that
KB-Sindy always sets to zero the 1st and 2nd order components.
Furthermore the 3rd and 5th order dynamics appear negligible compared to the 4th order component.
In fact, on average, the norm of the 4th order system component is almost one million times larger.
These outcomes may suggest to refine the estimate by using Sindy equipped only with 
4th order monomials, thus reducing the number of basis functions to 715.
Fig. \ref{NFIR3} shows the boxplot of the 100 prediction fits, which all appear close to $99\%$.
Although the number of basis functions is still considerable (even compared to the size of the data set),
Sindy's control of complexity is remarkable when only 4th order monomials are used.
This shows that the information about the sparsity pattern returned by KB-Sindy can be used also to
improve Sindy's performance. 

\clearpage

\bibliographystyle{plain}
\bibliography{biblio}  

\end{document}